%% file: seed_paper.tex
\documentclass[]{bytedance_seed}

\usepackage[toc,page,header]{appendix}
\usepackage{pifont}  % 用于 \ding{55} 叉号
\usepackage{booktabs}
\usepackage{xcolor} % 用于上色
\usepackage{rotating}
\usepackage{rotating}
\usepackage{adjustbox}
\usepackage{array,tabularx}
\usepackage{arydshln}
\usepackage{geometry}
\usepackage{makecell}
\usepackage{multirow}
\usepackage{float} % 放在导言区
\usepackage{amssymb}     % 提供 \checkmark
\usepackage{graphicx}
\usepackage{hyperref}
\hypersetup{
    colorlinks=true,
    linkcolor=blue,
    filecolor=magenta,      
    urlcolor=blue,
    citecolor=blue,
    breaklinks=true  % <--- 这个是关键！
}
\usepackage{url}
 
\newcolumntype{L}{>{\raggedright\arraybackslash}X}
\usepackage{minitoc}
\newcolumntype{C}{>{\centering\arraybackslash}X}
\newcolumntype{L}{>{\raggedright\arraybackslash}X}

\title{ MEF: A Systematic Evaluation Framework for Text-to-Image Models }
\affiliation[1]{ByteDance Seed}
\contribution[*]{Full author list in Contributions}
\abstract{
Rapid advances in text-to-image (T2I) generation have raised higher requirements for evaluation methodologies. Existing benchmarks center on objective capabilities and dimensions, but lack an application-scenario perspective, limiting external validity. Moreover, current evaluations typically rely on either ELO for overall ranking or MOS for dimension-specific scoring, yet both methods have inherent shortcomings and limited interpretability. Therefore, we introduce the Magic Evaluation Framework (MEF), a systematic and practical approach for evaluating T2I models. First, we propose a structured taxonomy encompassing user scenarios, elements, element compositions, and text expression forms to construct the Magic-Bench-377, which supports label-level assessment and ensures a balanced coverage of both user scenarios and capabilities. On this basis, we combine ELO and dimension-specific MOS to generate model rankings and fine-grained assessments respectively. This joint evaluation method further enables us to quantitatively analyze the contribution of each dimension to user satisfaction using multivariate logistic regression. By applying MEF to current T2I models, we obtain a leaderboard and key characteristics of the leading models. We release our evaluation framework and make Magic-Bench-377 fully open-source to advance research in the evaluation of visual generative models.
}
\date{\today}
\correspondence{\email{liushu@bytedance.com}}

\checkdata[Project Page]{\url{https://huggingface.co/datasets/ByteDance-Seed/MagicBench}}

\usepackage{CJK}
\begin{document}
\begin{CJK}{UTF8}{gbsn}
\maketitle
%不需要目录就注释掉 注意目录不要和第一页放在一块 要有\newpage
%\newpage
%\tableofcontents
%\newpage

\input{1_introduction}
\input{2_relatedwork}
\input{3_core_framework}

\input{4_benchmark_377}
\input{5_evaluation_metric}

\input{6_experiments}
\input{7_conclusion}
\input{8_future_work}

\clearpage

\beginappendix

\input{appendix}

\end{CJK}
\end{document}

%% file: 1_introduction.tex
\section{Introduction}
\begin{figure}
\centering
\includegraphics[width=0.9\linewidth]{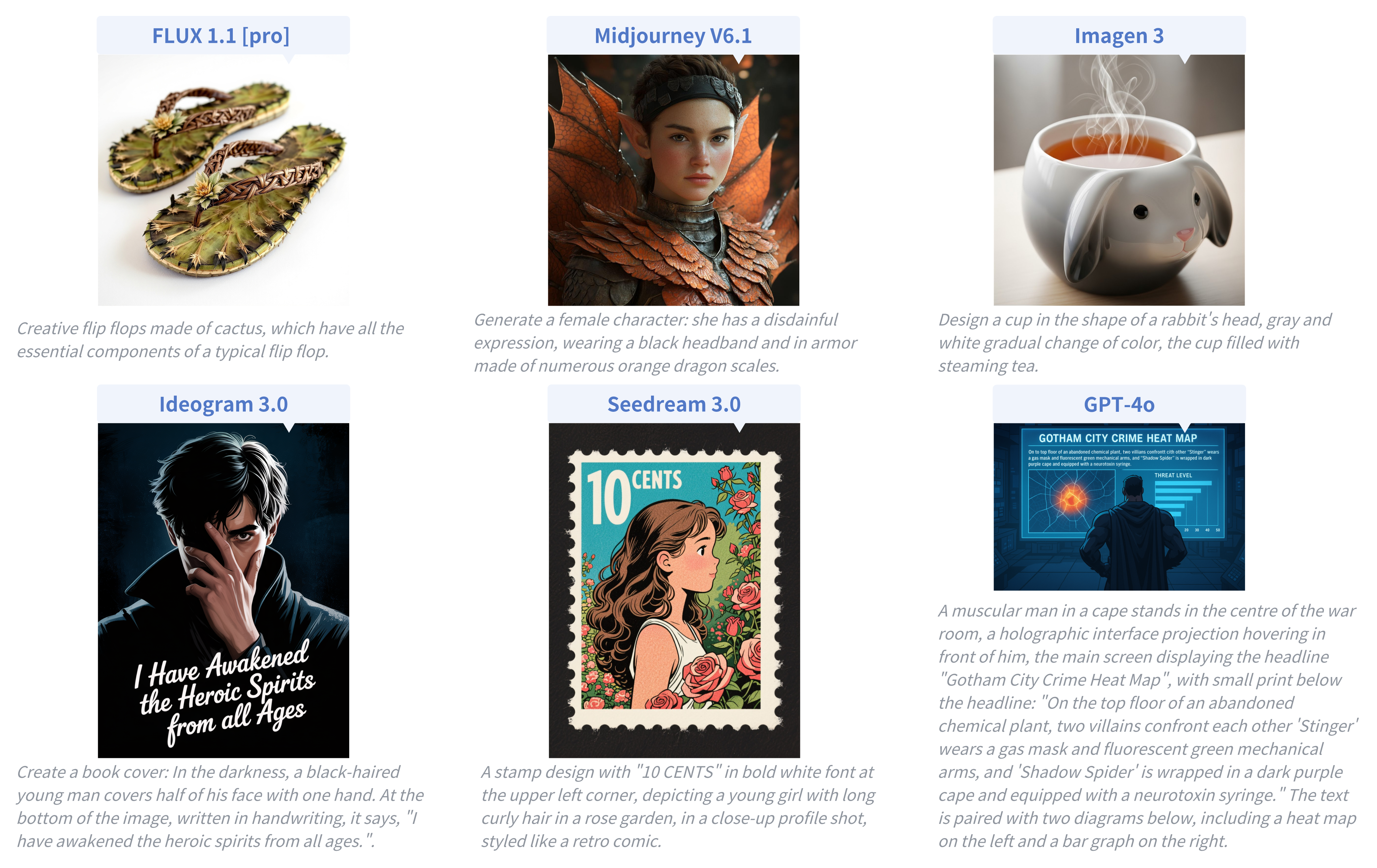}
\caption{\label{fig:overall}Example images are generated by T2I models. All prompts and images are sourced from Magic-Bench-377.}
\end{figure}

In recent years, the text-to-image (T2I) generation field has progressed rapidly through successive iterations. A new cohort of models, exemplified by Midjourney~\cite{Mun2025}, Ideogram~\cite{Bosheah2025}, and GPT-4o~\cite{Yan2025}, together with well-established baselines such as Stable Diffusion~\cite{rombach2022high} and DALL$\cdot$E~\cite{ramesh2021zero,ramesh2022hierarchical,Betker2023}, collectively represent current technological frontier. These models exhibit distinct comparative advantages across core capabilities (Fig.~\ref{fig:overall}): Midjourney series excels at artistic style rendering; Seedream~\cite{Seedream3.0} and Ideogram achieve breakthroughs in text rendering accuracy, excelling especially when handling complex textual prompts; and GPT-4o demonstrates strong performance in compositional and semantic understanding.

This diversification intensifies the demands on evaluation methodology. In addition to establishing a rigorous overall ranking, it is also important to conduct diagnostic assessments that efficiently and reliably identify performance differences in specific capabilities, such as quantity, style, and pronoun reference, as well as across representative application scenarios, including commercial design, artistic creation, and personal entertainment.

In contrast to the evaluation of large language models (LLMs) and vision language models (VLMs), assessing visual content generation introduces distinct challenges.

\begin{figure}[!htbp]
\centering
\includegraphics[width=0.85\linewidth]{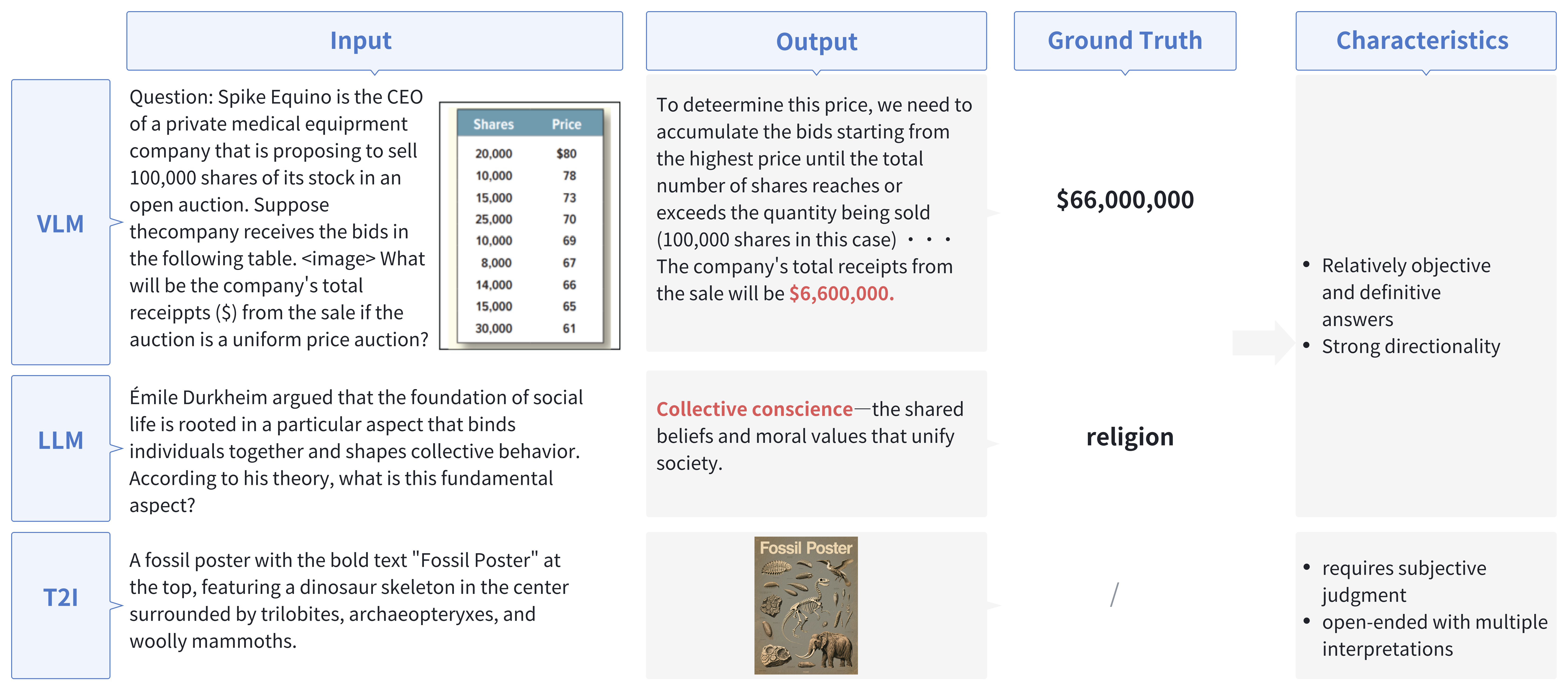}
\caption{\label{fig:eval_comparison}An example comparing inputs and outputs for LLM, VLM, and T2I models from a public benchmark~\cite{Yue2023}. LLMs and VLMs typically have standard answers, while T2Is are open-ended, allowing for diverse valid outputs.}
\end{figure}

\begin{figure}[!htbp]
\centering
\includegraphics[width=0.9\linewidth]{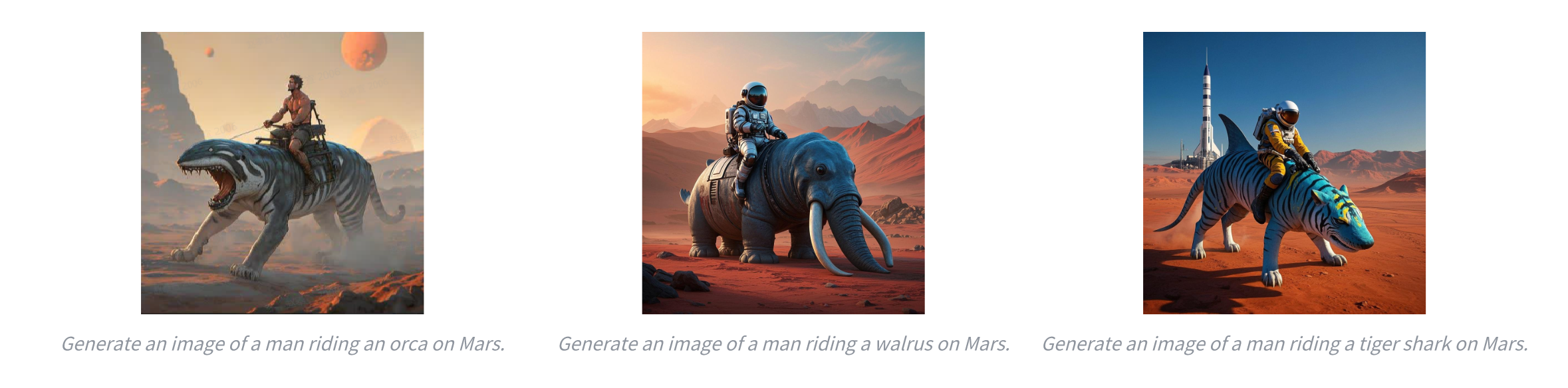}
\caption{\label{fig:ambiguous cases}Coupling between dimensions complicates evaluation. For instance,  these images fail to generate the correct subject, which create scoring ambiguity in Prompt Following and Structure Accuracy.}
\end{figure}

\begin{itemize}
    \item First, evaluation outcomes are highly sensitive to subjective factors (Fig.~\ref{fig:eval_comparison}), where scores are strongly influenced by the evaluators' visual preferences, cultural background, and professional expertise, especially in the aesthetic dimension. Even for more objective capabilities (e.g., object count accuracy, correctness of entity relations), composite tasks can introduce weighting ambiguities. For instance, within Prompt Following, when simultaneously assessing the numerical and spatial accuracy of a prompt like "three cats and two pillows on a sofa", the relative importance of each component often depends on the evaluator's individual judgment.
\end{itemize}
\begin{itemize}
    \item Second, there exist pronounced coupling effects across evaluation dimensions. For example, structural issues (e.g., disproportionate object scales) directly impact perceived aesthetics, while prompt-image mismatches (e.g., "orca" (Chinese in "$\mbox{虎鲸}$") being rendered as a tiger (Chinese in "$\mbox{虎}$") with a whale's mouth (Chinese in "$\mbox{鲸}$"), in Fig.~\ref{fig:ambiguous cases}) in turn affect judgments of structural plausibility. These imply that the automation-heavy approaches commonly used for LLMs~\cite{Chang2023,Laskar2024,Huang2024} and VLMs~\cite{Lee2024,Li2025,Xiong2024} are not well-suited for evaluating visual generative models. Instead, it is necessary to develop more flexible and domain-specific visual assessment systems.
\end{itemize}
Against this backdrop, we introduce the Magic Evaluation Framework (MEF), a systematic evaluation methodology with two principal contributions:
\begin{itemize}
    \item We construct a benchmark grounded in application scenarios and objective capability labels, enabling precise identification of model weaknesses in real world use and linking model capability to user experience across contexts. Each prompt is intentionally designed to interleave multiple capabilities, so that a single prompt assesses several skills simultaneously, approximating holistic performance in practical settings. At the same time, refined criteria reduce cross-dimensional interference, improving the fidelity of single-dimension assessments.
\end{itemize}
\begin{itemize}
    \item We integrate an ELO-based dynamic competitive ranking mechanism with a multi-dimensional Mean Opinion Score (MOS) protocol for absolute scoring. This cross-validated design provides reliable performance leaderboards and allows for detailed analysis of individual capabilities and the influence of different dimensions. At the same time, we implement an end-to-end quality assurance pipeline to ensure that, while reasonable subjective preferences are accommodated during evaluation, the outcomes remain objectively reliable.
\end{itemize}

As visual generative models advance, this rigorous evaluation system will be increasingly central to their assessment and development. By connecting model capability iteration with real-world application scenarios, the framework ensures that continual improvements are closely aligned with user experience.

%% file: 2_relatedwork.tex
\section{Related Work}
\subsection{Evaluation Methods}
%Evaluation methodologies for visual generative models have followed distinct trajectories and research priorities in academia and industry.
Evaluation methodologies for visual generative models have shown some differences between academia and industry, driven by their respective objectives and constraints.

In academia, efforts have predominantly focused on automated evaluation, evolving in two phases. Early approaches trained dedicated reward models~\cite{Kirstain2023}, leveraged mature computer vision backbones~\cite{Ghosh2023}, or incorporated LLM based assessors~\cite{Hu2023}.
%with the advent of multimodal models such as GPT-4o and Gemini that possess strong visual understanding capabilities, VLM-based direct evaluation or fine-tuned evaluation~\cite{Huang2025} has become the mainstream approach to automation.
With the emergence of multimodal models (e.g., GPT-4o, Gemini~\cite{team2023gemini}) possessing strong visual understanding, VLM based evaluation, either used directly or fine-tuned for task-specific objectives~\cite{Huang2025}, has become the prevailing paradigm for automation.

Overall, such methods perform reasonably well on Prompt Following but struggle to diagnose Structural Accuracy and Aesthetic Quality, which exhibit an obvious deviations from human preference.

In industry, evaluation practice prioritizes precise and fine-grained results and therefore employs labor-intensive, human-in-the-loop protocols. Two paradigms are prevalent:
\begin{itemize}
    \item The first centers on an ELO framework. Representative adopters include DALL$\cdot$E~\cite{Betker2023}, Imagen~\cite{Baldridge2024}, Stable Diffusion~\cite{Esser2024}, and third-party Arena platforms~\cite{Chiang2024}. Beyond producing a holistic leaderboard, the ELO mechanism is also extended to dimensional assessments---for example, DALL$\cdot$E reports ELO scores over Prompt Following, Style, and Coherence. Nevertheless, ELO rankings are inherently relative: pairwise comparisons provide only an order (Model A is better than Model B) but not absolute magnitude of differences, leading to a loss of fine-grained signal, particularly problematic under non-uniform distributions (e.g., strong wins on a small subset but slight losses on the majority).
\end{itemize}
\begin{itemize}
    \item The second centers on Mean Opinion Score (MOS). Representative adopters include Ketu~\cite{Kolors2024} and Wan~\cite{Wang2025}. For instance, Wan applies regression analysis to learn differentiated weights over 14 sub-dimensions and computes a weighted average to obtain a composite ranking. However, this still exhibits limited robustness as a ranking method: weights vary across user groups---professional designers, for example, typically emphasize aesthetic dimension more than general users; moreover, dimension importance weights evolve over time, making static weighting schemes ill-suited to capture shifting priorities.
\end{itemize}
\subsection{Benchmark for T2I Model Evaluation}
Current benchmarks for evaluating T2I models have made progress in comprehensively assessing model capabilities. However, they still suffer from issues such as unsystematic and incomplete capability categorization, as well as insufficient alignment with real application scenarios (in Table.~\ref{tab:transposed_grouped}). For instance, DrawBench~\cite{Saharia2022}, with its 200 prompts, covers 11 scattered categories such as color and quantity. Yet, its capability labels are presented in an enumerated manner (e.g., placing "rare words" and "DALL$\cdot$E dataset" side by side), which may hinder holistic assessment of model capabilities. PartiPrompt~\cite{Yu2022} extends the scale to 1,632 prompts and introduces a "12 capability categories $\times$ 3 difficulty levels" evaluation matrix. Nonetheless, its category design also tends toward simple enumeration (e.g., treating abstract content, world knowledge, and human figures as parallel categories), which also limits comprehensive capability assessment.
%Benchmarks focusing on compositional ability, such as T2I-CompBench++~\cite{Huang2025b} and GenEval~\cite{Ghosh2024}, are restricted in scope: the former to four capabilities including attribute binding and object relations, and the latter to object-centric properties such as color and quantity.
Benchmarks targeting compositionality, such as T2I-CompBench++~\cite{Huang2025} and GenEval~\cite{Ghosh2023}, remain narrow in scope: the former covers four capabilities (including attribute binding and object relations), whereas the latter emphasizes object-centric properties (e.g., color and quantity).
Both fail to adequately cover complex semantics or specialized forms of expression. GenAI-Bench~\cite{Li2024} introduces a two-tiered capability hierarchy---"basic" (entity/attribute) and "advanced" (logic/comparison). However, its 1,600 prompts are not tied to user scenarios, and its advanced capabilities do not account for competencies about textual expression form. DEsignBench~\cite{Lin2023} contributes 215 prompts targeting design-related scenario, but these are limited to visual design contexts and associated technical abilities, excluding domains such as film, art, and entertainment, thereby limiting its applicability in multi-scenario model selection. Other benchmarks such as WISE~\cite{Niu2025} (semantic understanding), CUBE~\cite{Kannen2024} (world knowledge), Commonsense~\cite{Wu2024} (commonsense reasoning), and GradBias~\cite{DInca2025} (social bias) each focus on a single dimension of capability. While valuable for probing specific weaknesses, they are not suitable for evaluating overall model performance.

\begin{table}[htbp]
\centering
\small
\begin{adjustbox}{max width=\textwidth}
% 注意：我把 p{2.5cm} 改成了 c，讓 makecell 自己決定寬度，這樣更靈活
% 如果你一定要固定寬度，也可以保留 p{2.5cm}
\begin{tabular}{ll*{7}{c}} 
\toprule
\multirow{2}{*}{} & \multirow{2}{*}{} 
    & \multicolumn{7}{c}{\textbf{Benchmarks}} \\
\cmidrule(lr){3-9}
& & \makecell[b]{PartiPrompt\\\cite{Yu2022}} % <-- 加上 [b]
   & \makecell[b]{DrawBench\\\cite{Saharia2022}} % <-- 加上 [b]
   & \makecell[b]{T2I-CompBench++\\\cite{Huang2025}} % <-- 加上 [b]
   & \makecell[b]{GenEval\\\cite{Ghosh2023}} % <-- 加上 [b]
   & \makecell[b]{GenAI-Bench\\\cite{Li2024}} % <-- 加上 [b]
   & \makecell[b]{DESignBench\\\cite{Lin2023}} % <-- 加上 [b]
   & \makecell[b]{Magic-Bench-377\\\vphantom{\cite{Lin2023}}} \\
\midrule
\multirow{5}{*}{\textbf{Application Scenarios}} 
& Film              & \textcolor{red}{\ding{55}} & \textcolor{red}{\ding{55}} & \textcolor{red}{\ding{55}} & \textcolor{red}{\ding{55}} & \textcolor{red}{\ding{55}} & \textcolor{red}{\ding{55}} & \checkmark \\
& Art               & \textcolor{red}{\ding{55}} & \textcolor{red}{\ding{55}} & \textcolor{red}{\ding{55}} & \textcolor{red}{\ding{55}} & \textcolor{red}{\ding{55}} & \textcolor{red}{\ding{55}} & \checkmark \\
& Entertainment     & \textcolor{red}{\ding{55}} & \textcolor{red}{\ding{55}} & \textcolor{red}{\ding{55}} & \textcolor{red}{\ding{55}} & \textcolor{red}{\ding{55}} & \textcolor{red}{\ding{55}} & \checkmark \\
& Aesthetic Design  & \textcolor{red}{\ding{55}} & \textcolor{red}{\ding{55}} & \textcolor{red}{\ding{55}} & \textcolor{red}{\ding{55}} & \textcolor{red}{\ding{55}} & \checkmark                 & \checkmark \\
& Functional Design & \textcolor{red}{\ding{55}} & \textcolor{red}{\ding{55}} & \textcolor{red}{\ding{55}} & \textcolor{red}{\ding{55}} & \textcolor{red}{\ding{55}} & \checkmark                 & \checkmark \\
\midrule
\multirow{14}{*}{\textbf{Objective Capabilities}} 
& Entity                        & \checkmark & \checkmark & \checkmark & \checkmark & \checkmark & \textcolor{red}{\ding{55}} & \checkmark \\
& Quantity                      & \checkmark & \checkmark & \checkmark & \checkmark & \textcolor{red}{\ding{55}} & \textcolor{red}{\ding{55}} & \checkmark \\
& Attribute                     & \checkmark & \checkmark & \checkmark & \checkmark & \checkmark & \textcolor{red}{\ding{55}} & \checkmark \\
& Relation                      & \checkmark & \checkmark & \checkmark & \checkmark & \textcolor{red}{\ding{55}} & \textcolor{red}{\ding{55}} & \checkmark \\
& Action/State                  & \checkmark & \checkmark & \checkmark & \checkmark & \textcolor{red}{\ding{55}} & \textcolor{red}{\ding{55}} & \checkmark \\
& Style                         & \checkmark & \textcolor{red}{\ding{55}} & \textcolor{red}{\ding{55}} & \textcolor{red}{\ding{55}} & \textcolor{red}{\ding{55}} & \textcolor{red}{\ding{55}} & \checkmark \\
& Aesthetic                     & \checkmark & \textcolor{red}{\ding{55}} & \textcolor{red}{\ding{55}} & \textcolor{red}{\ding{55}} & \checkmark & \textcolor{red}{\ding{55}} & \checkmark \\
& Atmosphere                    & \textcolor{red}{\ding{55}} & \textcolor{red}{\ding{55}} & \textcolor{red}{\ding{55}} & \textcolor{red}{\ding{55}} & \textcolor{red}{\ding{55}} & \checkmark & \checkmark \\
& Multi-Entity Feature Matching & \checkmark & \checkmark & \checkmark & \checkmark & \checkmark & \textcolor{red}{\ding{55}} & \checkmark \\
& Layout \& Typography          & \textcolor{red}{\ding{55}} & \checkmark & \checkmark & \checkmark & \checkmark & \checkmark & \checkmark \\
& Anti-Realism                  & \checkmark & \checkmark & \checkmark & \checkmark & \checkmark & \textcolor{red}{\ding{55}} & \checkmark \\
& Negation                      & \textcolor{red}{\ding{55}} & \textcolor{red}{\ding{55}} & \textcolor{red}{\ding{55}} & \textcolor{red}{\ding{55}} & \checkmark & \textcolor{red}{\ding{55}} & \checkmark \\
& Pronoun Reference             & \textcolor{red}{\ding{55}} & \textcolor{red}{\ding{55}} & \textcolor{red}{\ding{55}} & \textcolor{red}{\ding{55}} & \checkmark & \textcolor{red}{\ding{55}} & \checkmark \\
& Consistency                   & \checkmark & \checkmark & \checkmark & \textcolor{red}{\ding{55}} & \checkmark & \textcolor{red}{\ding{55}} & \checkmark \\
\bottomrule
\end{tabular}
\end{adjustbox}
\caption{Coverage differences between Magic-Bench-377 and existing T2I evaluation benchmarks in terms of application scenarios and objective capabilities. Magic-Bench-377 provides broader coverage of both core capabilities and application domains. The methodology for calculating label coverage is provided in Appendix~\ref{appendix:A}.}
\label{tab:transposed_grouped}
\end{table}

From this, we can identify the key challenges faced by current evaluation methodologies:

\begin{itemize}
    \item The separation between ELO and MOS evaluation paradigms prevents existing systems from satisfying, at once, the dual requirements of credible ranking and fine-grained diagnostics.
\end{itemize}

\begin{itemize}
    \item There is a lack of diagnostic tooling calibrated to concrete application scenarios, giving rise to a persistent "technology to experience" gap. A recurring question from model developers is: how much business value does an improvement in technical metrics actually generate?
\end{itemize}

    Together, these challenges pose obstacles to the reliability of visual generative model evaluation and may affect the field's ongoing development. To address this, it is necessary to establish a more comprehensive theoretical framework for evaluation, enhancing both scientific rigor and practical relevance.
    
    In summary, existing efforts have not resolved the disconnection between model capability and real-world application. To bridge this gap, we introduce a new evaluation framework.

%\subsection{Hello World}

%% file: 3_core_framework.tex
\section{Core Framework}
\begin{figure}[!htbp]
\centering
\includegraphics[width=0.9\linewidth]{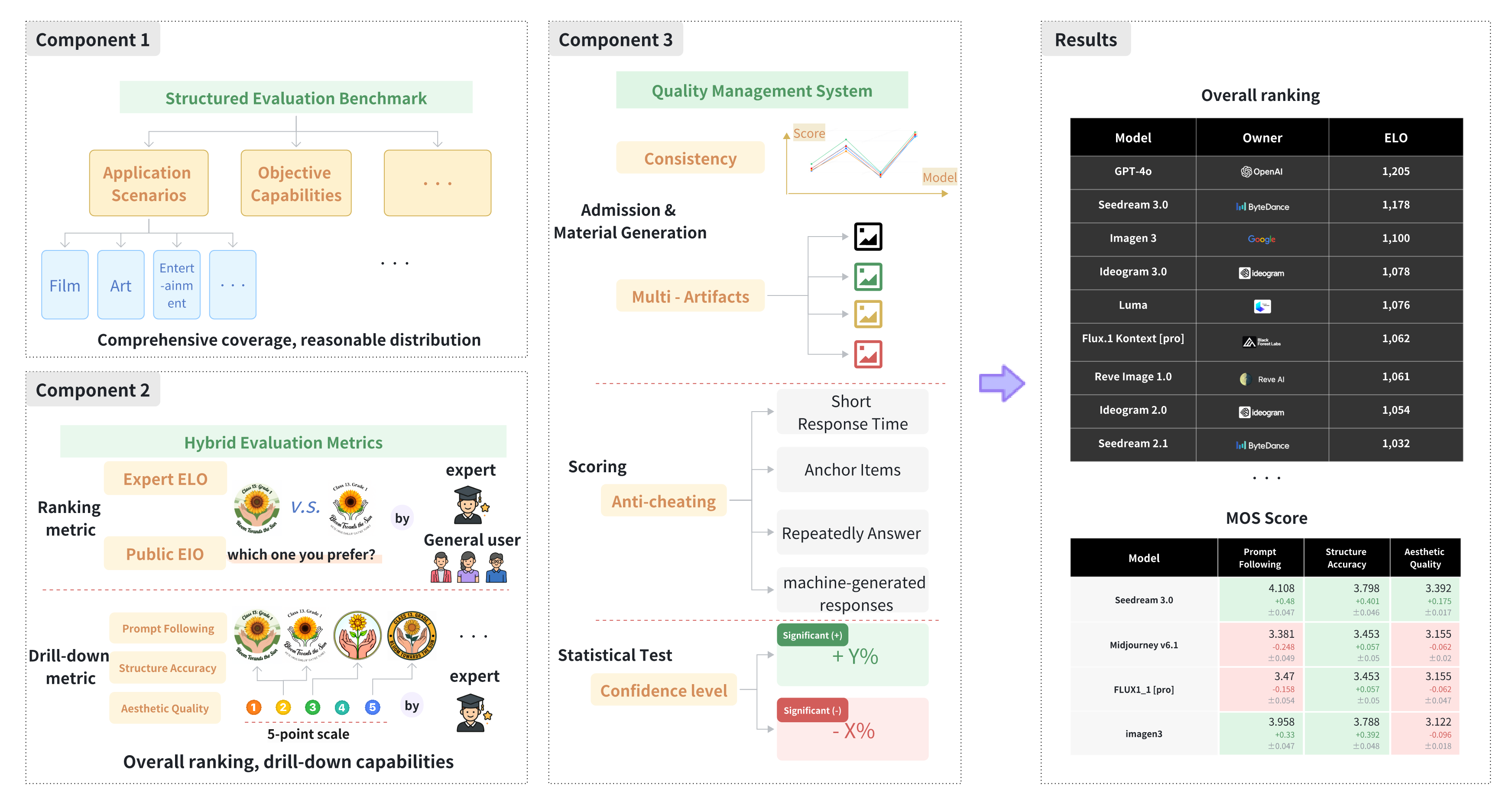}
\caption{\label{fig:MEF}Benchmarks, hybrid evaluation metrics, and a quality management system constitute the three core pillars of the Magic Evaluation Framework.}
\end{figure}

Our core contribution is an evaluation system for T2I generation models---the Magic Evaluation Framework (MEF, in Fig.~\ref{fig:MEF}). The framework comprises three mutually reinforcing components: (1) a structured benchmark, (2) hybrid evaluation metrics, and (3) an end-to-end quality management system. As a result, MEF can efficiently quantify the performance of both objective capabilities and applications, and provide guidance for iterative model improvement.

\begin{table}[htbp]
    \centering
    \small 
    \resizebox{0.8\textwidth}{!}{
    \begin{tabular}{>{\centering\arraybackslash}p{3.5cm} p{4cm} p{5cm} p{2cm}}
        \toprule
        & & & \textbf{Percentage} \\
        \midrule
        \multirow{5}{*}{\textbf{Application Scenarios}} & & \hspace{1em}Film & 20\% \\
        & & \hspace{1em}Art & 21\% \\
        & & \hspace{1em}Entertainment & 12\% \\
        & & \hspace{1em}Aesthetic Design & 25\% \\
        & & \hspace{1em}Functional Design & 22\% \\
        \midrule
        \multirow{13}{*}{\textbf{Objective Capability}} 
        & \multirow{4}{*}{Entity Description} 
            & \hspace{1em}Quantity & 4\% \\
        &   & \hspace{1em}Attribute & 9\% \\
        &   & \hspace{1em}Relation & 12\% \\
        &   & \hspace{1em}Action / State & 10\% \\
        \cmidrule(lr){2-4}
        & \multirow{3}{*}{Visual Description} 
            & \hspace{1em}Style & 64\% \\
        &   & \hspace{1em}Aesthetic & 35\% \\
        &   & \hspace{1em}Atmosphere & 6\% \\
        \cmidrule(lr){2-4}
        & \multirow{3}{*}{Element Composition} 
            & \hspace{1em}Multi-Entity Feature Matching & 8\% \\
        &   & \hspace{1em}Layout \& Typography & 5\% \\
        &   & \hspace{1em}Anti-Realism & 8\% \\
        \cmidrule(lr){2-4}
        & \multirow{3}{*}{Textual Expression Form} 
            & \hspace{1em}Negation & 2\% \\
        &   & \hspace{1em}Pronoun Reference & 1\% \\
        &   & \hspace{1em}Consistency & 2\% \\
        \bottomrule
    \end{tabular}
    }
    \caption{Distribution of objective capability and application scenario labels in the Magic-Bench-377 benchmark. Each evaluation item is annotated with multiple objective capability labels and a single application scenario label. The rationale behind the label distribution is detailed in the section on benchmark construction methodology.~\hyperref[sec:benchmark_construction]{(Section~\ref{sec:benchmark_construction})}}
    \label{tab:objective_distribution}
\end{table}

\begin{itemize}
    \item \textbf{Benchmark.} The benchmark consists of 377 text prompts, each annotated with dual labels---hundreds of  objective capabilities (e.g., entity-relation understanding, numerical understanding) and 5 application scenarios (e.g., film, art). Based on our prior practice and user research, we develop a systematic taxonomy for text-to-image generation and determine the distribution within the benchmark (Table~\ref{tab:objective_distribution}), balancing high-frequency, high-value, and high-difficulty cases. Consequently, the evaluation can reflect real-world model performance and also distinguish the specific characteristics of different models. 
    \item \textbf{Hybrid evaluation metrics.} We adopt an integrated paradigm that combines ELO with MOS. The ELO mechanism aggregates tens of thousands of anonymized head-to-head matches to produce a dynamically updated leaderboard that provides a holistic view of model ranking. The MOS mechanism assesses three core aspects---Prompt Following, Structural Accuracy, and Aesthetic Quality---on a refined 1--5 scale, providing dimension-level assessments. For each model, we generate four test images per prompt, which are then scored back-to-back by an expert panel. This complementary methods allow us to further estimate the contribution of each dimension to user satisfaction through logistic regression.
    \item \textbf{End-to-end quality management system.} We institute quality controls throughout the pipeline to ensure the reliability of both ELO and MOS. During expert evaluation, we enforce a strict three-phase protocol: Magic-Bench-377 provides broad capability and scenario coverage; evaluators must possess solid grasp of key points to score accurately, and must avoid cross-dimensional interference among the three base dimensions (Prompt Following, Structural Accuracy, and Aesthetic Quality). Accordingly, all evaluators who undergo standardized training must pass qualification and consistency tests before participation, and are periodically re-certified. During execution, we deploy multiple quality-control (QC) strategies: automatically insert 5\% anchor items for real-time monitoring and conduct manual rechecks on 25\% of samples by QC specialists. The final MOS report includes raw data and comprehensive statistical analyses, including 95\% confidence intervals (95\% CI) and significance testing (p < 0.01). In this way, every step from implementation to reporting, is supported by traceable, reproducible quality controls, ensuring high statistical power for our conclusions.
\end{itemize}

\begin{comment}
\begin{table}[htbp]
\centering
\small
% 定义第一列宽度固定为3cm，第二列自动换行
\begin{tabularx}{\textwidth}{>{\raggedright\arraybackslash}p{4cm} L}
\toprule
\textbf{Basic Concept} & \textbf{Definition} \\
\midrule
Metrics & Multiple independent evaluation metrics for assessing the performance of text-to-image generation models, such as Prompt Following, Structural Accuracy, and Aesthetics. \\
Objective Capabilities & The capability items that a text-to-image generation model should possess, categorized based on the mapping relationship between textual prompts and visual elements. \\
Application Scenarios & Application types categorized in practical scenarios based on core service objectives and differing requirements, such as art and film. \\
Test Points & The current model still has shortcomings in certain objective capabilities during the generation process, such as relational-size relationships and pronoun reference. The test points represent a subset of objective capabilities. \\
\bottomrule
\end{tabularx}
\caption{Basic definitions of key elements in the MEF}
\label{tab:basic_definitions}
\end{table}
\end{comment}

%% file: 4_benchmark_377.tex
\section{Magic-Bench-377}

\subsection{Taxonomy}
\begin{figure}[!htbp] % 或 [H]
    \centering
    \includegraphics[width=0.8\textwidth]{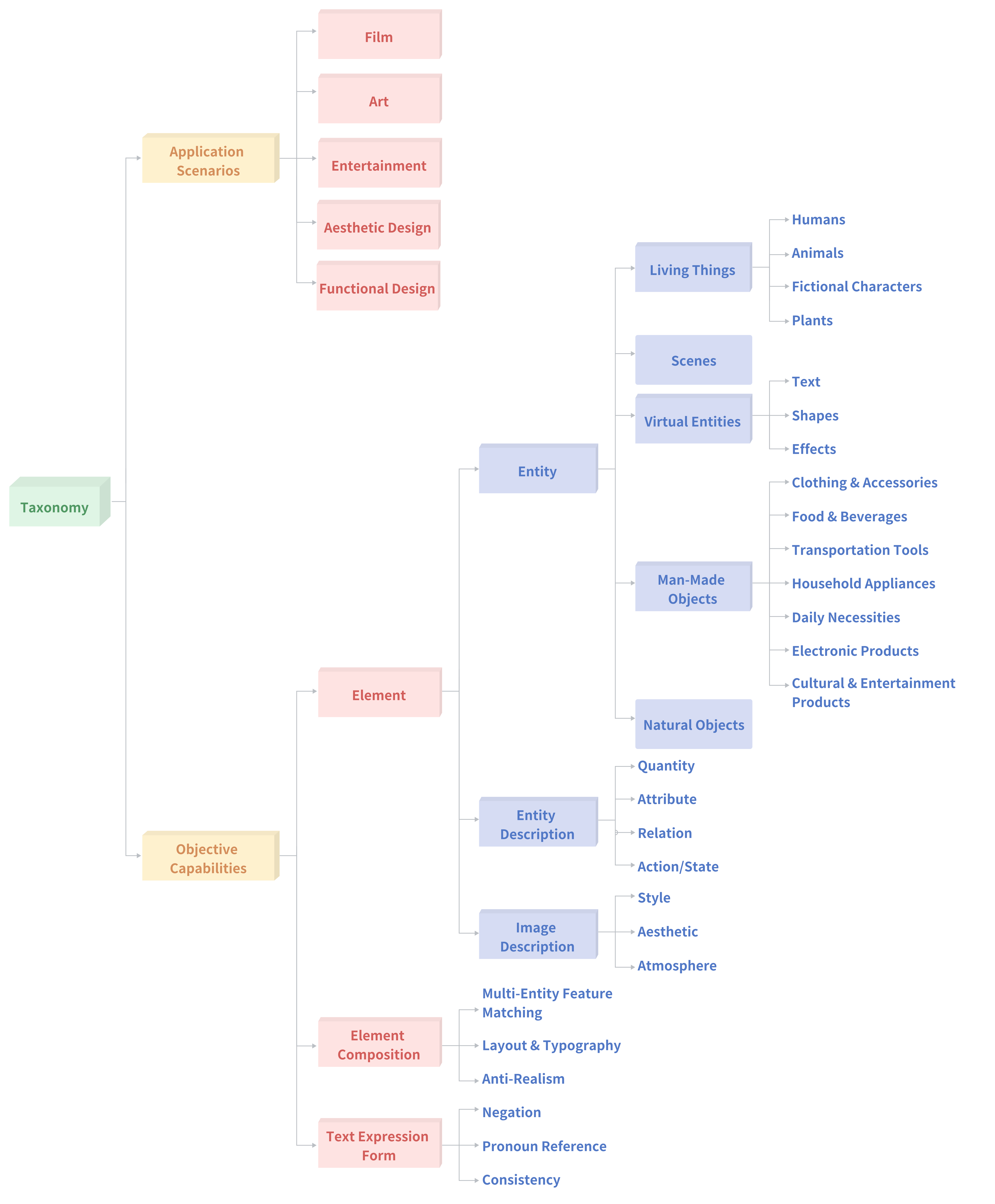}
    \caption{Taxonomy for T2I Models. The taxonomy encompasses two primary branches: application scenarios and objective capabilities. These labels are designed to be extensible, with some further subdivided into finer categories in practical use.}
    \label{fig:taxonomy}
\end{figure}

To facilitate the effective construction of evaluation prompt sets, we develop a taxonomy that systematically captures the core capabilities and application scenarios of T2I models. This taxonomy is established based on prior evaluation experience, user and expert interviews, and sector-wide studies, ensuring that the evaluation datasets are both comprehensive and well-structured (Fig.~\ref{fig:taxonomy}).

\textbf{Objective Capabilities.} We categorize model capabilities by integrating textual prompts and visual elements, dividing them into three mutually exclusive categories:

\begin{itemize}
    \item \textbf{Element.} Refers to visual elements or information that can be expressed by a single semantic unit, typically a word (e.g., owl, running). Depending on the types of visual element expressed, these are further subdivided into Entity, Entity Description, and Image Description.
    \item \textbf{Element Composition.} Refers to visual elements or information arising from the combination of multiple elements (e.g., an imagined creature possessing feature X of entity A, and feature Y of entity B). Based on prior evaluation findings, we emphasize three composition types where current models still exhibit notable limitations: Multi-Entity Feature Matching (multiple entities of the same type with different features), Layout $\&$ Typography (spatial composition of multiple entities), Anti-Realism (combinations that violate real world laws).
    \item \textbf{Text Expression Form.} Refers to semantic units not directly pointing to visual elements but testing a model's understanding and reasoning over special forms of expression. Examples include Negation (e.g., "a living room without a coffee table") and Pronoun Reference (e.g., "the bear lies on the ground, the cub lies beside it").
\end{itemize}

Definitions and example prompts for each label are provided in Appendix~\ref{appendix:B}, with the "Consistency" label adapted from the GenAI benchmark taxonomy~\cite{Li2024}.

\textbf{Application Scenarios.} To account for the diversity of real world user demands, we divide application scenarios for T2I models into five categories: Film, Art, Entertainment, Aesthetic Design, and Functional Design. Definitions and example prompts are detailed in Appendix~\ref{appendix:B}.

\subsection{Benchmark Construction Methodology}

\begin{figure}[!htbp]
\centering
\includegraphics[width=0.85\linewidth]{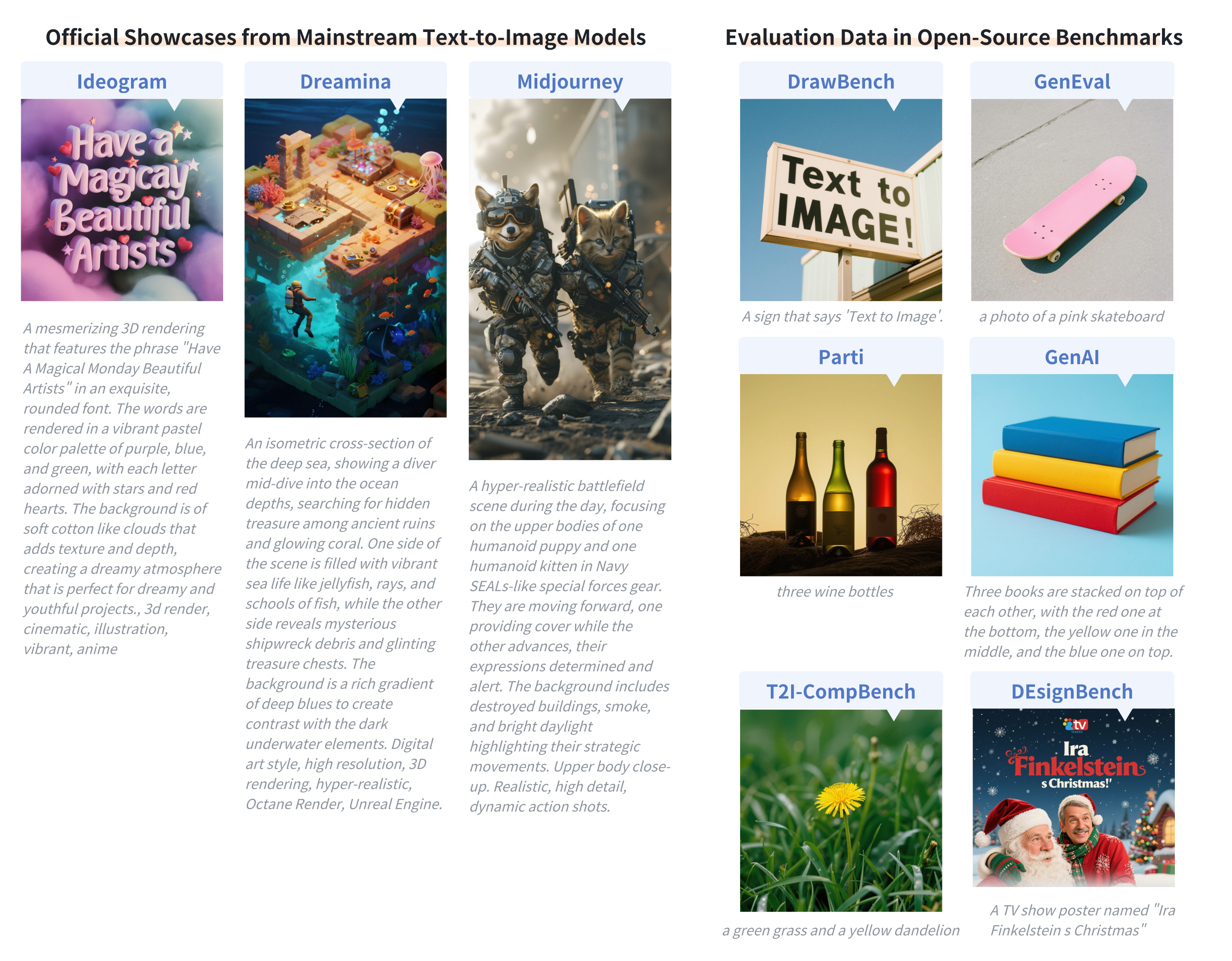}
\caption{\label{fig:showcase}Examples from official showcases of mainstream models and open-source benchmark evaluation datasets. Source: Official Website~\cite{Showcase_sources}. Open-source benchmarks tend to be concise and focus on single capability, which differs from practical scenarios where user prompts often describe more diverse scenes and complex entities.}
\end{figure}

As illustrated in Fig.~\ref{fig:showcase}, real-world prompts often specify complete scenes that require multiple capabilities, whereas existing benchmarks (e.g., Parti~\cite{Yu2022}, DrawBench~\cite{Saharia2022}) typically assess a single capability per prompt.
We observe that interactions among multiple test points\footnote{Test point is defined based on previously identified weaknesses of T2I models in specific aspects and is a subset of objective capabilities.} can negatively impact the performance on a given test point compared to evaluating it in isolation. 

For instance, when numerical and size relations are presented together in a single prompt, the model exhibits an obvious decrease in performance on the numerical aspect, relative to prompts focused solely on numerical relations (Appendix.~\ref{appendix:C}). This cross-capability interference can produce a mismatch between user experience and evaluation results, and may overstate performance in realistic use. 
To better reflect the user perspective, Magic-Bench-377 embeds multiple capabilities within a single prompt. Concretely, each prompt in Magic-Bench-377 is annotated with 1--4 objective capability labels and 1 application scenario label.

\subsection{Benchmark Construction Procedure}\label{sec:benchmark_construction}
\begin{figure}[!htbp]
\centering
\includegraphics[width=1\linewidth]{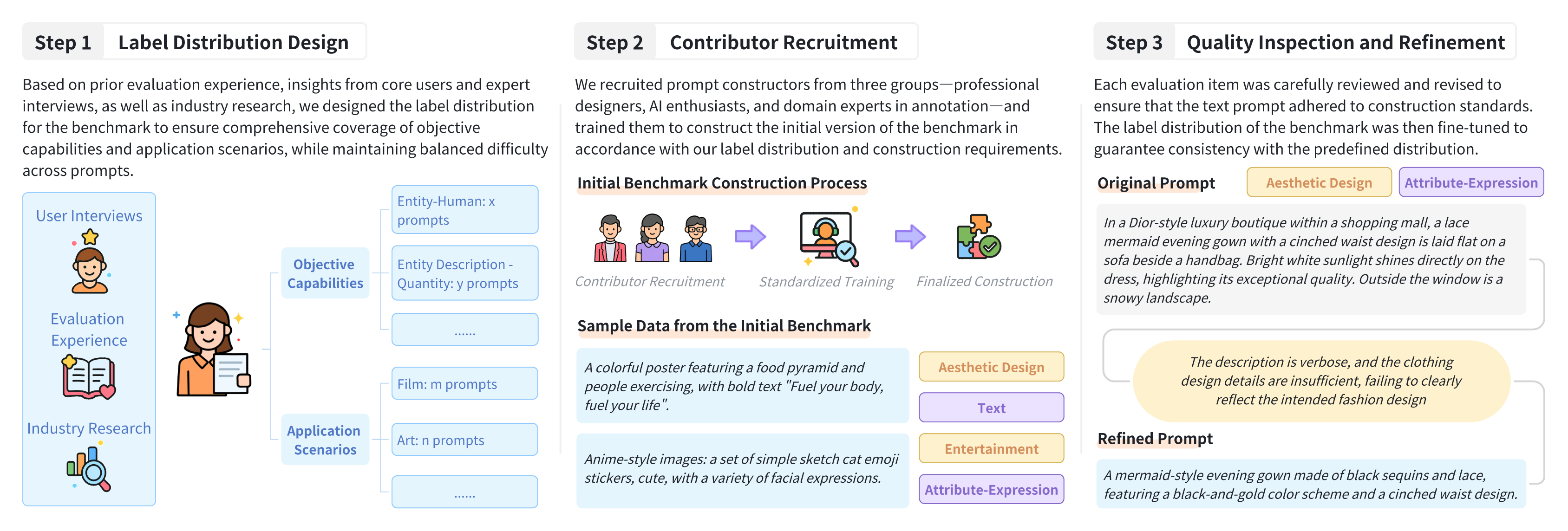}
\caption{\label{fig:benchmark_construction}Benchmark Construction Procedure}
\end{figure}

The construction steps of Magic-Bench-377 are as follows (Fig.~\ref{fig:benchmark_construction}):

\begin{itemize}
    \item\textbf{Step 1: Designing the label distribution.} Drawing upon previous evaluation experience, we systematically identified objective capabilities that remain challenging for current models (hereafter referred to "model challenges"). By defining capabilities and scenarios based on current user needs and our long-term value judgments, and by utilizing "model challenges" to regulate difficulty levels, we constructed a benchmark that ensures both balanced and comprehensive coverage.
    \item\textbf{Step 2: Recruiting contributors to construct evaluation prompts.} To better align with real world requirements, we recruited a diverse set of contributors with prompt writing experience. Specifically, we engaged (1) professional designers, (2) AI enthusiasts with extensive experience using visual generation tools, and (3) domain experts with rich annotation experience. Prior to prompt construction, all contributors received standardized training to ensure adherence to the predefined prompt construction standards (see Appendix~\ref{appendix:C}). Prompt-design task were assigned to the best suited contributor groups according to professional backgrounds, yielding a high quality initial benchmark. 
    
    During data construction, we specified the application scenario and capability labels for each prompt, and then selected corresponding textual expressions and visual elements. These elements were subsequently composed in a natural manner, ensuring that the resulting prompts are representative of reasonable user requirements. For instance, when assessing Negation, "there are no scallions in the noodles" serves as a better prompt than "there is no elephant in the room".
    \item\textbf{Step 3: Quality inspection and prompt refinement.} Once the initial benchmark was constructed, we reviewed and revised each prompt to ensure that its quality and overall distribution were consistent with our expectations.
\end{itemize}

%% file: 5_evaluation_metric.tex
\section{Evaluation Metric}
\subsection{ELO Score}
\subsubsection{System Overview}
During ELO matchups, we determine the winner based on the image's overall quality, which essentially aligns with the three dimensions used in MOS evaluation: Prompt Following, Structural Accuracy, and Aesthetic Quality. If Image A is better on at least one dimension and no worse on the others, A wins over B. When trade-offs exist, evaluators rely on a holistic assessment that considers the severity of errors across the dimensions.

We designed and implemented a head-to-head platform~\cite{AIGCArena2024} tailored for visual evaluation, featuring periodic leaderboard update. At the global level, a composite leaderboard derived from the full set of user data can be calculated, while at the individual level, personalized leaderboards are provided. The leaderboard presents the following key statistics: (1) ELO score; (2) 95$\%$ confidence intervals; (3) current model rank; (4) cumulative matches played; and (5) win rate.

In the evaluation process, the platform employs a double-blind design: two anonymized model outputs are presented side by side in the user interface. Evaluators then select one of four outcomes: left wins / right wins / both good / both bad. 

\textbf{ELO score estimation.} We follow the Chatbot Arena framework~\cite{Chiang2024}, construct a win-rate matrix under the Bradley-Terry (BT) model and estimate coefficients via maximum likelihood. Here, we define win rate of model A against other models as $\frac{W_A + 0.5\, T_A}{N_A}$, where $W_A$ denotes the number of wins by A, $T_A$ the number of ties, and $N_A$ the total number of matches involving A. Model ELO ratings are then obtained by a linear transformation of the BT coefficients.
\begin{equation}
    P(H_t = 1) = \frac{1}{1 + e^{\xi'_m - \xi_m}},
\end{equation}
Let $H_t \in \{0,1\}$ be a Bernoulli random variable representing the user's preference outcome: $H_t = 1$ if the user selects model $m$; $H_t = 0$ if the user selects model $m'$. $\xi_m, \xi_{m'}$ denote the BT coefficients of models $m$ and $m'$, respectively. Additionally, we fix the baseline model's ELO rating at 1,000.

Furthermore, we are able to compute the ELO score for each prompt, providing additional insight into prompt-level contributions. Specifically, to compute ELO for $prompt_i$, we retain the full match schedule but replace the outcomes of all matches not involving  $prompt_i$ with draws. By iterating this procedure over all prompts, we obtain each prompt's ELO contribution:

\begin{equation}
  \mathrm{ELO}_{\text{model}} \approx \sum_{i=1}^{377} \mathrm{ELO}_{\mathrm{prompt}_i},\quad \text{err}<5\%
\end{equation}

\textbf{Expert mode and public mode.} In expert mode, evaluations are performed by a carefully selected, well-trained group of expert panel, typically fewer than 20 individuals, ensuring consistency and accuracy. In public mode, we broadly recruit participants from diverse sectors such as design, media, and film/TV, as well as AI enthusiasts. The user base exceeds one thousand, providing a more representative reflection of general user preferences. The two modes are completely independent in their matching workflows, computation pipelines, and final leaderboard generation. Comparative analysis of leaderboards from expert and public modes reveals that while most model rankings are aligned, certain divergences exist. These differences can serve to calibrate expert standards, ensuring they better reflect the preferences and needs of general users.

\subsubsection{Ensuring Scientific Validity}

\textbf{Confidence interval estimation and baseline model selection. }We use bootstrap resampling to obtain the empirical distribution of the BT coefficients~\cite{Chiang2024}: repeatedly resample and refit the model, then compute 95\% confidence intervals using the percentile method. Under this setup, since the baseline model's ELO score is fixed at 1,000 (and thus has no confidence interval), we select a lower-ranked model as the baseline to minimize potential bias in the ranking of top models.

\textbf{Efficient Match Scheduling Mechanism.} Rather than relying on purely random pairings, we aim to improve match efficiency and facilitate faster convergence of ELO scores on the leaderboard. To achieve this, we sample model pairs with fewer historical matches at a higher probability, so newly launched models have a greater chance of being matched in order to accelerate convergence. We also prioritize model pairs whose win-loss probabilities are close to 50\%, as these matchups are most effective in narrowing their ELO confidence intervals.

Operationally, we begin by selecting a model pair based on historical match data in accordance with the aforementioned principles. Next, we uniformly sample a prompt from Magic-Bench-377, and randomly select one output from each model's multiple generations. Finally, these outputs are presented to evaluators.

\textbf{Minimum number of matches.} To ensure leaderboard stability, we set a minimum participation threshold for listed models. The threshold is the maximum of two criteria: the confidence interval width for each model must narrow to 20 ELO points or less, and the ELO change induced by a single match must not exceed 3 points. In our benchmark protocol, this typically requires models to complete more than 4,000 head-to-head matches to appear on the leaderboard.

\textbf{Anti-cheating mechanisms. }Casual interactions by some users, whether for trial or novelty, can introduce noise that affects ranking accuracy. By deploying anti-cheating mechanisms, we achieve a 30\% reduction in the ELO confidence interval width across models, which substantially enhances the ranking reliability.

\begin{figure}[!htbp]
\centering
\includegraphics[width=0.8\linewidth]{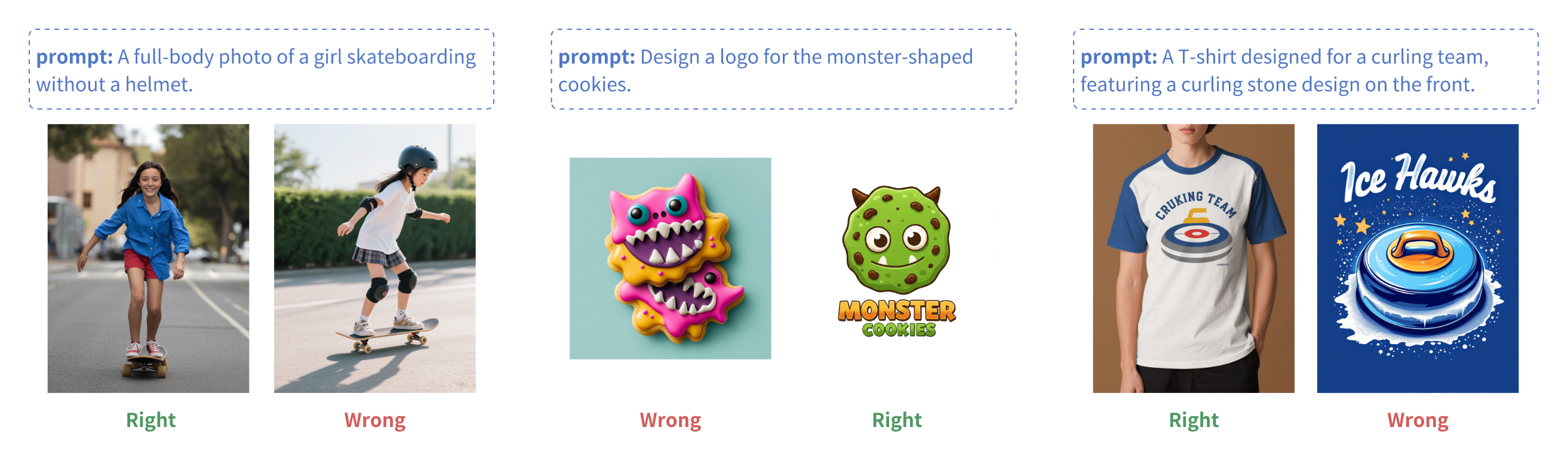}
\caption{\label{fig:anchor}Typical examples of anchor pairs. These anchor prompts are selected through iterative rounds of evaluation. On these prompts, participants' choices exhibit a high degree of consistency.}
\end{figure}

Specifically, we mainly use temporal behavior analysis and anchor item tests to filter out cheating users. Temporal analysis identifies abnormal responses by detecting unusually fast submissions and excessive answer repetition. For example, completion speeds more than three standard deviations faster than the user average, or repeatedly selecting the same option dozens of times, are considered strong indicators of cheating. In addition, anchor item tests (Fig.~\ref{fig:anchor}) randomly insert 5\% of pairs with verified outcomes, which are pairs exhibiting pronounced quality differences. Users who consistently fail to identify the higher quality image in these items are flagged as invalid evaluators.

In expert mode, additional qualification prescreening and dynamic monitoring are implemented. Expert assessor are required to pass a simulated matchup test, achieving at least 85\% agreement with the expert cohort (Kappa > 0.8) to qualify. During formal evaluation, 25\% of assessments are randomly sampled for re-auditing, and experts not meeting quality standards will be temporarily suspended from scoring.

\subsection{MOS Score}
\subsubsection{System Overview}

\begin{figure}[!htbp]
\centering
\includegraphics[width=0.9\linewidth]{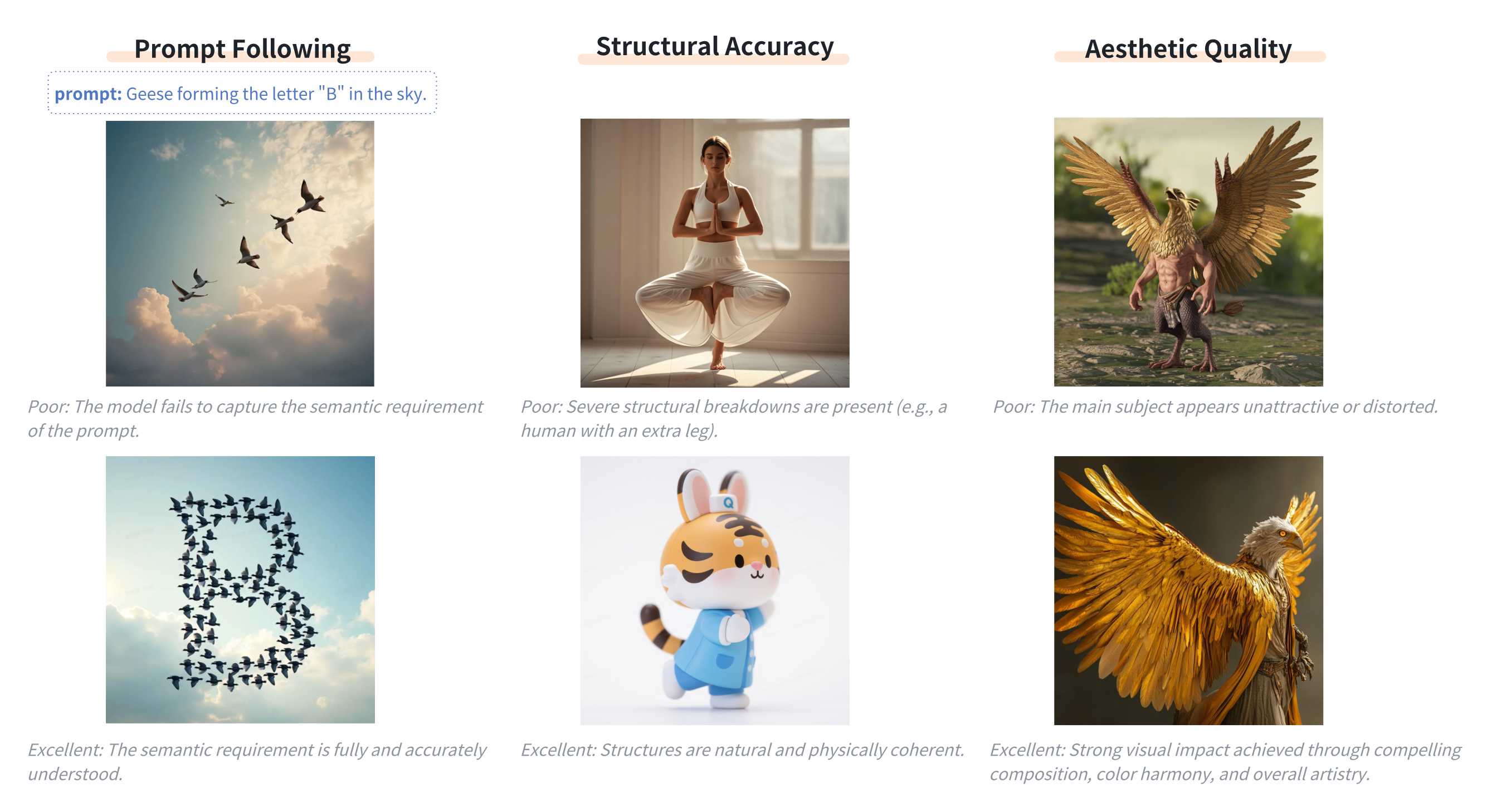}
\caption{\label{fig:cases of metrics}Illustrative evaluation guidelines for the three core MOS dimensions.}
\end{figure}

In addition to the ELO based ranking mechanism, MEF incorporates a Mean Opinion Score (MOS) system to further assess model performance. This system conducts absolute evaluations across three dimensions, Prompt Following, Structural Accuracy, and Aesthetic Quality, each scored on a 1–-5 scale by the same expert panel as in ELO expert mode. As shown in Fig.~\ref{fig:cases of metrics}:

\begin{itemize}
    \item Prompt Following assesses the degree of alignment between the generated image and the input textual prompt at the semantic level.
\end{itemize}
\begin{itemize}
    \item Structural Accuracy evaluates the completeness of entities within the image, as well as whether their structures and interrelations conform to real world cognitive expectations and physical commonsense.
\end{itemize}
\begin{itemize}
    \item Aesthetic Quality considers the visual appeal of the generated image from an artistic perspective, focusing on style rendering, color harmony, composition, and the interplay of light and shadow.
\end{itemize}

\subsubsection{Ensuring Scientific Validity}

\textbf{Multi-sample scoring.} To reduce the impact of generation stochasticity, the benchmark adopts a multi-sample evaluation mechanism. Specifically, for each test prompt, we generate four independent samples per model to more accurately reflect a model's average generative capability. Experiments show that when evaluating a single sample only, the probability of significant variation across different generations from the same model can reach 30\% (self-comparison test). Using a four-sample averaged evaluation strategy reduces this probability to below 5\% ($p < 0.01$), markedly improving evaluation stability.
%Procedurally, we implement a dual-expert review.
%Each expert can simultaneously inspect all models' outputs for the same prompt, enabling cross-comparison and holistic judgment.
For each prompt, two experts independently score all models' outputs side-by-side, enabling cross-comparison judgment.
This design effectively mitigates biases that might arise from isolated evaluations.

\begin{figure}[!htbp]
\centering
\includegraphics[width=0.8\linewidth]{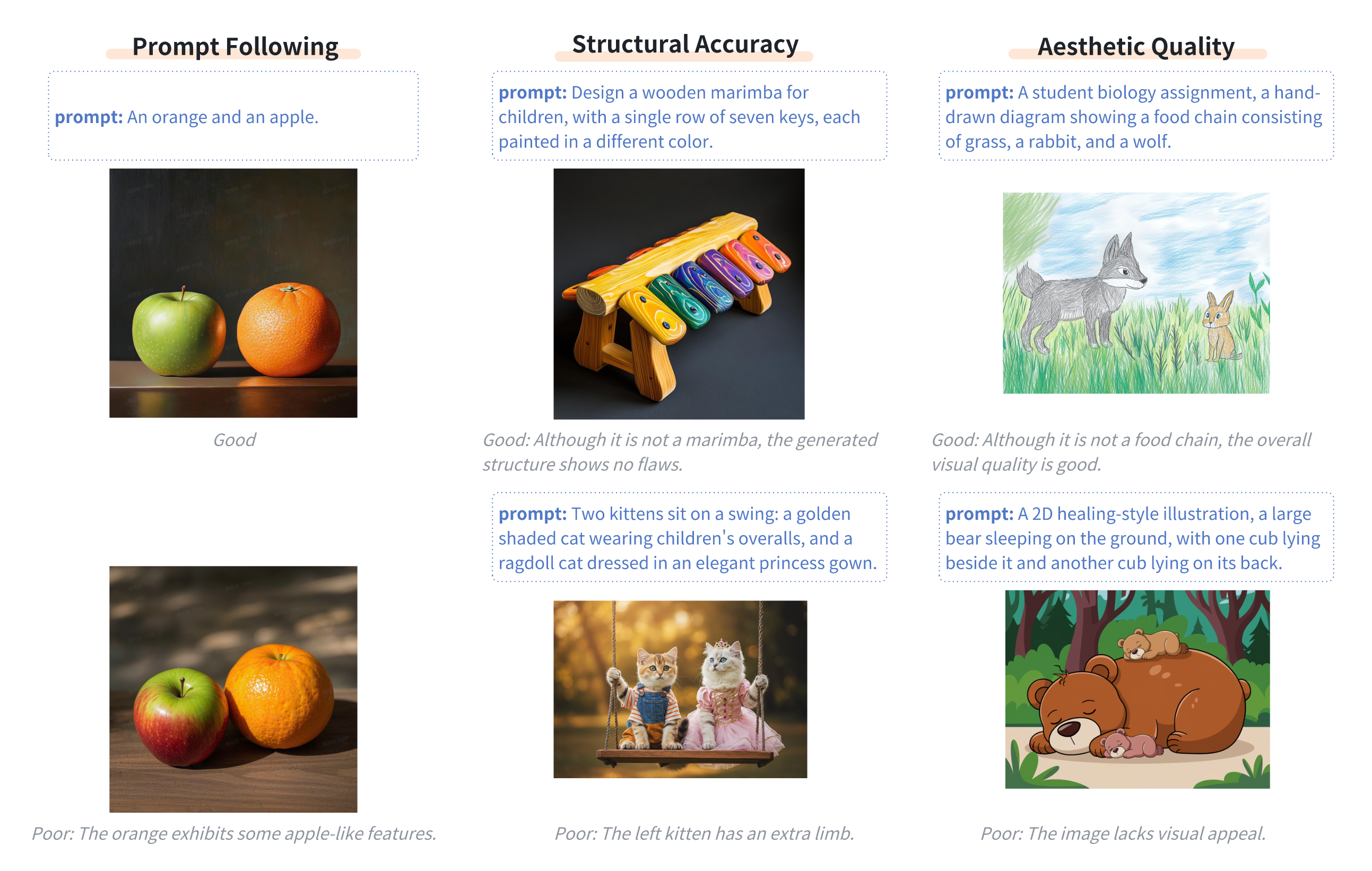}
\caption{\label{fig:reduce_coupling}Example cases demonstrating how evaluation standards are defined to reduce coupling effects across the three core dimensions.}
\end{figure}

\textbf{Evaluation standards training.} To minimize interference caused by coupling effects among the three dimensions (in Fig.~\ref{fig:reduce_coupling}), we establish clear guidelines and conduct expert training to ensure these standards are fully understood and correctly applied.

When assessing Prompt Following, evaluators should focus exclusively on the alignment between the generated image and the input prompt, without being influenced by structural or aesthetic quality. Deductions are permitted only if the structural errors are so severe that they directly affect semantic interpretation. Similarly, when evaluating Structural Accuracy or Aesthetic Quality, evaluators should concentrate solely on the core criteria of the respective dimension, disregarding other factors unless severe issues in another dimension directly impact their judgment. Detailed scoring criteria are provided in Appendix~\ref{appendix:E}.

\begin{figure}
\centering
\includegraphics[width=0.25\linewidth]{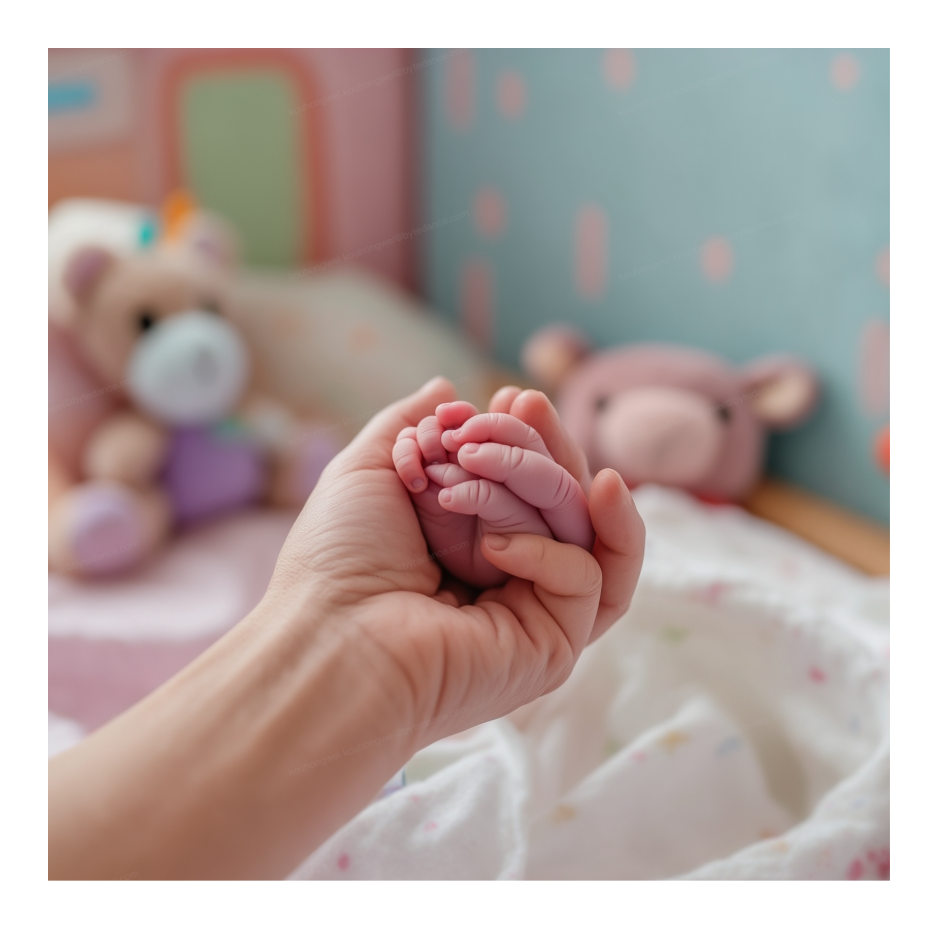}
\caption{\label{fig:structural performance}Severe weaknesses in structural performance tend to undermine the perceived aesthetic quality of an image.}
\end{figure}

To validate the mutual independence across the three MOS dimensions, we compute Pearson correlation coefficients for each assessor. The statistical analysis shows that all inter-dimensional correlations are below 0.3 ($r < 0.3$, Table~\ref{tab:correlation_coefficients}), indicating that the dimensions are nearly decoupled. The correlation between Structural Accuracy and Aesthetic Quality is slightly higher, as severe structural failures can notably impact perceived aesthetics in certain scenarios (in Fig.~\ref{fig:structural performance}).    

\begin{table}[!htbp]
\centering
\small
\begin{adjustbox}{max width=0.75\textwidth}
\begin{tabular}{>{\bfseries}p{8cm} cccc}
\toprule
& Expert A & Expert B & Expert C & Expert D \\
\midrule
\textbf{Prompt Following \& Structural Accuracy} & 0.172 & 0.142 & 0.141 & 0.123 \\
\textbf{Prompt Following \& Aesthetic Quality} & 0.111 & 0.153 & 0.124 & 0.091 \\
\textbf{Structural Accuracy \& Aesthetic Quality} & 0.259 & 0.272 & 0.27  & 0.275 \\
\bottomrule
\end{tabular}
\end{adjustbox}
\caption{The Pearson correlation coefficients between MOS dimensions, illustrated using data from four experts.}
\label{tab:correlation_coefficients}
\end{table}

\noindent
\textbf{Confidence estimation.} Traditional confidence estimation rests on a random sampling assumption and uses $\frac{\sigma^2}{n}$ (where $\sigma^2$ is the sample variance and $n$ is the number of items). This approach requires that per-item generation outcomes are independent and identically distributed (i.i.d.). However, contemporary industry benchmarks are manually curated collections of specialized prompts and thus do not satisfy the random sampling assumption.
We therefore propose an improved confidence estimation framework. Suppose the benchmark contains $n$ prompts, each evaluated $k$ independent times. Let $a_{i,j}$ denote the $j^{th}$ evaluation outcome for prompt $i$. The per-dimension mean score is computed as follows:
\begin{equation}
\mathrm{M} = \frac{1}{n} \sum_{i=1}^{n} \left( \frac{1}{k} \sum_{j=1}^{k} a_{i,j} \right)
\end{equation}
Under the i.i.d. assumption within each prompt, the variance estimator of the MOS score is:
\begin{equation}
\mathrm{var}(\mathrm{M}) = \frac{1}{n^2 \times k} \sum_{i=1}^{n} \mathrm{var}(a_i)
\end{equation}
Based on our prior evaluation studies, a MOS difference greater than 0.1 between two models on a specific dimension typically suggests a statistically significant difference.

%% file: 6_experiments.tex
\section{Experimental Result}
\subsection{Experimental setup}

This study evaluates mainstream T2I models on Magic-Bench-377 using the MEF framework. For each prompt, every model generates four images, which are used uniformly for both ELO and MOS evaluations. Specifically, the results for public and expert ELO evaluations are derived from 106,287 and 62,736 valid matches, respectively.
\subsection{ELO Score}

\textbf{Expert Mode Evaluation Results.} As shown in Table~\ref{tab:ELO_scores}, GPT-4o and Seedream 3.0 are first-tier models, with GPT-4o leading across all scenarios and Seedream 3.0 performing comparably in Art, Functional Design, and Aesthetic Design. Imagen 3, Ideogram 3.0, and Luma fall into the middle tier with relatively balanced performance, whereas Recraft V3 Raw and Midjourney V6.1 underperform in ELO, primarily due to notable weaknesses in certain dimensions.
\begin{table}[!htbp]
\centering
\small
\setlength{\tabcolsep}{2pt} % 调整列间距
\begin{adjustbox}{max width=\textwidth}
\begin{tabular}{l
  cc cc cc cc cc cc
}
\toprule
\multirow{2}{*}{\textbf{Model~\cite{Showcase_sources}}} 
& \multicolumn{2}{c}{\textbf{Overall}} 
& \multicolumn{2}{c}{\textbf{Art}} 
& \multicolumn{2}{c}{\textbf{Functional Design}} 
& \multicolumn{2}{c}{\textbf{Aesthetic Design}} 
& \multicolumn{2}{c}{\textbf{Film}} 
& \multicolumn{2}{c}{\textbf{Entertainment}} \\
\cmidrule(lr){2-3} \cmidrule(lr){4-5} \cmidrule(lr){6-7} \cmidrule(lr){8-9} \cmidrule(lr){10-11} \cmidrule(lr){12-13}
& Rank & ELO Score & Rank & ELO Score & Rank & ELO Score & Rank & ELO Score & Rank & ELO Score & Rank & ELO Score \\
\midrule
GPT-4o & 1 & \bf{1,205} & 1 & \bf{1,162} & 1 & \bf{1,157} & 1 & \bf{1,125} & 1 & \bf{1159} & 1 & \bf{1173} \\
Seedream 3.0 & 2 & 1,178 & 2 & 1,147 & 2 & 1,108 & 2 & 1,117 & 2 & 1,139 & 2 & 1,115 \\
Imagen 3 & 7 & 1,100 & 6 & 1,077 & 7 & 1,038 & 7 & 1,038 & 7 & 1,033 & 8 & 1,052 \\
Ideogram 3.0 & 7 & 1,078 & 13 & 988 & 9 & 1,026 & 9 & 1,020 & 12 & 963 & 9 & 1,030 \\
Luma & 8 & 1,076 & 8 & 1,033 & 11 & 1,003 & 8 & 1,028 & 8 & 1,030 & 10 & 1,030 \\
Flux.1 Kontext [pro] & 8 & 1,062 & 11 & 994 & 10 & 1,019 & 10 & 1,018 & 10 & 990 & 11 & 1018 \\
Reve Image 1.0 & 9 & 1,061 & 12 & 992 & 8 & 1,033 & 12 & 994 & 11 & 981 & 7 & 1,061 \\
Ideogram 2.0 & 9 & 1,054 & 10 & 1,000 & 12 & 1,000 & 11 & 1,000 & 9 & 1,000 & 13 & 1,000 \\
Seedream 2.1 & 13 & 1,032 & 15 & 982 & 14 & 985 & 13 & 972 & 13 & 959 & 14 & 998 \\
Recraft V3 Raw & 13 & 1,021 & 17 & 958 & 13 & 989 & 14 & 970 & 14 & 947 & 15 & 984 \\
Flux 1.1 [pro] & 14 & 1,011 & 16 & 960 & 15 & 962 & 15 & 937 & 15 & 945 & 12 & 1,004 \\
Midjourney V6.1 & 16 & 1,000 & 9 & 1,003 & 16 & 936 & 16 & 931 & 17 & 910 & 16 & 964 \\
\bottomrule
\end{tabular}
\end{adjustbox}
\caption{ELO scores under expert mode. The rankings appear non-consecutive, since we display only the key models and preserve their original positions from the full leaderboard.}
\label{tab:ELO_scores}
\end{table}

\textbf{Public Mode Evaluation Results.} As shown in Table~\ref{tab:ELO_public}, Seedream 3.0 achieves the best overall performance and ranks first, followed closely by GPT-4o. At the scenario level, Seedream 3.0 consistently outperforms in Art, Functional Design, Aesthetic Design, and Film, while GPT-4o achieves its best results in Functional Design, Film, and Entertainment. The tier of remaining models is consistent with the expert-mode ELO results.
\begin{table}[htbp]
\centering
\small
\setlength{\tabcolsep}{2pt} % 调整列间距
\begin{adjustbox}{max width=\textwidth}
\begin{tabular}{l
  cc cc cc cc cc cc
}
\toprule
\multirow{2}{*}{\textbf{Model~\cite{Showcase_sources}}} 
& \multicolumn{2}{c}{\textbf{Overall}} 
& \multicolumn{2}{c}{\textbf{Art}} 
& \multicolumn{2}{c}{\textbf{Functional Design}} 
& \multicolumn{2}{c}{\textbf{Aesthetic Design}} 
& \multicolumn{2}{c}{\textbf{Film}} 
& \multicolumn{2}{c}{\textbf{Entertainment}} \\
\cmidrule(lr){2-3} \cmidrule(lr){4-5} \cmidrule(lr){6-7} \cmidrule(lr){8-9} \cmidrule(lr){10-11} \cmidrule(lr){12-13}
& Rank & ELO Score & Rank & ELO Score & Rank & ELO Score & Rank & ELO Score & Rank & ELO Score & Rank & ELO Score \\
\midrule
GPT-4o                 & 3  & 1,067 & 2  & 1,064 & 1  & 1,057 & 3  & 1,052 & 1  & 1,108 & 1  & 1,056 \\
Seedream 3.0           & 1  & \bf{1,084} & 1  & \bf{1,083} & 1  & \bf{1,071} & 1  & \bf{1,076} & 1  & \bf{1,112} & 1  & \bf{1,079} \\
Imagen 3               & 5  & 1,031 & 5  & 1,039 & 1  & 1,028 & 5  & 1,024 & 5  & 1,043 & 2  & 1,035 \\
Ideogram 3.0           & 8  & 1,007 & 11 & 986  & 4  & 1,020 & 9  & 998  & 7  & 1,015 & 4  & 1,019 \\
Luma                   & 8  & 1,017 & 7  & 1,014 & 9  & 996  & 5  & 1,012 & 5  & 1,025 & 3  & 1,007 \\
Flux.1 Kontext [pro]   & 7  & 1,019 & 8  & 1,008 & 5  & 1,012 & 5  & 1,012 & 5  & 1,024 & 2  & 1,035 \\
Reve Image 1.0         & 14 & 986  & 14 & 982  & 12 & 980  & 9  & 971  & 12 & 998  & 6  & 1,009 \\
Ideogram 2.0           & 15 & 984  & 16 & 975  & 13 & 970  & 12 & 986  & 11 & 1,019 & 12 & 982  \\
Seedream 2.1           & 8  & 1,017 & 5  & 1,014 & 6  & 1,007 & 5  & 1,017 & 5  & 1,029 & 5  & 1,014 \\
Recraft V3 Raw         & 11 & 997  & 7  & 1,004 & 9  & 994  & 10 & 991  & 9  & 995  & 9  & 989  \\
Flux 1.1 [pro]          & 13 & 992  & 7  & 1,010 & 14 & 967  & 12 & 982  & 10 & 1,005 & 10 & 1,006 \\
Midjourney V6.1        & 13 & 989  & 5  & 1,023 & 14 & 953  & 12 & 988  & 11 & 995  & 9  & 987  \\
\bottomrule
\end{tabular}
\end{adjustbox}
\caption{ELO scores under public mode, with the same set of models as in expert mode.}
\label{tab:ELO_public}
\end{table}

\textbf{Comparison of Expert and Public Perspectives.} Comparing the two rankings, both Seedream 3.0 and GPT-4o consistently appear at the top, indicating stable high performance across different user groups. The key difference lies in ELO score dispersion. The expert mode leaderboard shows greater separation, with nearly 200 ELO points between the top and bottom models, whereas the public mode leaderboard is much tighter, with only a 78 ELO point gap. This pattern suggests that experts exhibit greater sensitivity to subtle differences and, as a result, are more likely to select clear winners and losers among models, while general users tend to produce more ties.
\begin{table}[!htbp]
  \centering
  \begin{adjustbox}{max width=0.9\textwidth}
  \begin{tabular}{llccc}
    \toprule
    \multirow{2}{*}{\textbf{Model}} & \multirow{2}{*}{\textbf{Creator}} & \multicolumn{3}{c}{\textbf{ELO Score}} \\
    \cmidrule(lr){3-5}
    & & Magic-Bench-377 Expert & Magic-Bench-377 Public & Artificial Analysis \\
    \midrule
    GPT-4o               & OpenAI            & \bf{1,205} & 1,067 & 1,164 \\
    Seedream 3.0       & ByteDance         & 1,178 & \bf{1,084} & \bf{1,166} \\
    Imagen 3             & Google            & 1,100 & 1,031 & 1,097 \\
    Ideogram 3.0         & Ideogram AI       & 1,078 & 1,007 & 1,093 \\
    Luma                 & Dream Machine     & 1,076 & 1,017 & 1,040 \\
    FLUX.1 Kontext [pro] & Black Forest Labs & 1,062 & 1,019 & 1,101 \\
    Reve Image 1.0       & Reve AI           & 1,061 & 986   & 1,090 \\
    Ideogram 2.0         & Ideogram AI       & 1,054 & 984   & 1,045 \\
    Recraft V3 Raw       & Recraft           & 1,021 & 997   & 1,114 \\
    Flux 1.1 [pro]        & Black Forest Labs & 1,011 & 992   & 1,083 \\
    Midjourney V6.1      & Midjourney        & 1,000 & 989   & 1,047 \\
    \bottomrule
  \end{tabular}
  \end{adjustbox}
  \caption{Comparison of leaderboard results between Artificial Analysis and Magic-Bench-377.}
  \label{tab:comparison_aa_bench377}
\end{table}

\textbf{Comparison with public leaderboards.} We compare MEF ELO scores with the third-party public leaderboard Artificial Analysis (AA)\cite{AA_leaderboard} in Table~\ref{tab:comparison_aa_bench377}. The results show strong consistency between MEF and AA. Seedream 2.1 (the previous version of Seedream series) did not participate in AA, while all other models have corresponding ELO scores on AA. Statistically, the Pearson correlation of ELO scores across these models is close to 0.8, reflecting strong agreement. Achieving this level of correlation with just 377 prompts demonstrates the diverse coverage and strong representational power of Magic-Bench-377. While some models still show differences in ELO performance, these are mainly due to discrepancies in prompt distribution and evaluator composition between the two evaluation systems. 

\subsection{MOS Score}
Based on prior evaluation experience, we selected six representative models for MOS evaluation: Seedream 3.0, GPT-4o, Imagen 3, Ideogram 3.0, Flux 1.1 [pro], and Midjourney V6.1. Table~\ref{tab:model_scores} presents the evaluation results of these six models across the three fundamental dimensions---Prompt Following, Structural Accuracy, and Aesthetic Quality. For each dimension, both overall performance and results for five application scenarios are presented.

\begin{table}[htbp]
\centering
\begin{adjustbox}{max width=0.95\textwidth}
\begin{tabular}{%
  >{\raggedright\arraybackslash}p{2.5cm}
  >{\raggedright\arraybackslash}p{3cm}
  c c c c c c
}
\toprule
\textbf{Model} & & \textbf{Seedream 3.0} & \textbf{GPT-4o} & \textbf{Imagen 3} & \textbf{Ideogram 3.0} & \textbf{Flux 1.1 [pro]} & \textbf{Midjourney V6.1} \\
\midrule
\multirow{6}{=}{\textbf{Prompt Following}}
& Overall             & 4.23 & \bf{4.52} & 3.96 & 3.91 & 3.48 & 3.36 \\
& Aesthetic Design    & 4.17 & \bf{4.42} & 3.92 & 3.80 & 3.34 & 3.25 \\
& Functional Design   & 4.19 & \bf{4.54} & 3.89 & 4.06 & 3.59 & 3.27 \\
& Film                & 4.28 & \bf{4.51} & 3.85 & 3.70 & 3.32 & 3.04 \\
& Art                 & 4.28 & \bf{4.53} & 4.08 & 3.85 & 3.48 & 3.69 \\
& Entertainment       & 4.28 & \bf{4.74} & 4.14 & 4.12 & 3.80 & 3.67 \\
\midrule
\multirow{6}{=}{\textbf{Structural Accuracy}} 
& Overall             & 3.87 & \bf{4.23} & 3.79 & 3.71 & 3.41 & 3.31 \\
& Aesthetic Design    & 3.94 & \bf{4.36} & 3.79 & 3.90 & 3.47 & 3.28 \\
& Functional Design   & 3.78 & \bf{4.17} & 3.82 & 3.69 & 3.25 & 3.15 \\
& Film                & 3.69 & \bf{4.07} & 3.55 & 3.43 & 3.28 & 3.23 \\
& Art                 & 4.00 & \bf{4.18} & 3.92 & 3.68 & 3.53 & 3.61 \\
& Entertainment       & 3.98 & \bf{4.34} & 3.88 & 3.84 & 3.55 & 3.30 \\
\midrule
\multirow{6}{=}{\textbf{Aesthetic Quality}} 
& Overall             & \bf{3.39} & 3.17 & 3.12 & 3.07 & 3.15 & \bf{3.38} \\
& Aesthetic Design    & \bf{3.38} & 3.07 & 3.10 & 3.07 & 3.11 & 3.31 \\
& Functional Design   & \bf{3.34} & 3.13 & 3.12 & 3.10 & 3.09 & \bf{3.31} \\
& Film                & \bf{3.43} & 3.28 & 3.13 & 3.03 & 3.19 & \bf{3.44} \\
& Art                 & 3.42 & 3.22 & 3.14 & 3.04 & 3.21 & \bf{3.47} \\
& Entertainment       & 3.35 & 3.18 & 3.11 & 3.10 & 3.19 & \bf{3.41} \\
\bottomrule
\end{tabular}
\end{adjustbox}
\caption{Scores of Prompt Following, Structural Accuracy, and Aesthetic Quality across five application scenarios, with boldface indicating the top values in each row.}
\label{tab:model_scores}
\end{table}

As shown in Table~\ref{tab:model_scores}, mainstream models score above 3 across three dimensions, suggesting that the generated images generally do not exhibit significant deficiencies. However, performance varies by dimension and application scenarios, and no single model leads across the board.

{ \bf Characteristics of Mainstream Models.} By analyzing ELO scores, MOS scores, and the taxonomy labels of the prompts, we observe divergent characteristics among the six models:
\begin{itemize}
    \item GPT-4o: Strong semantic and structural performance. GPT-4o excels in Prompt Following (4.52) and structural accuracy (4.23), consistently outperforming all other models across scenarios, and is thus especially well-suited for professional contexts requiring semantic complexity and structural rigor, such as Entertainment and Functional Design. While its Aesthetic Quality score (3.17) is lower than that of Seedream 3.0 (3.39) and Midjourney V6.1 (3.38), this does not significantly affect its overall performance, as it still achieves the highest ELO scores.
\end{itemize}
\begin{itemize}
    \item Seedream 3.0: Seedream 3.0 shows balanced performance across the three dimensions without notable shortcomings, and its Aesthetic Quality score (3.39) is the highest among all models. It also demonstrates balanced advantages across scenarios, as reflected in both ELO and MOS scores. These advantages are especially pronounced in Art and Entertainment, which emphasize visual appeal and creativity.
\end{itemize}
\begin{itemize}
    \item Midjourney V6.1: Aesthetic specialist. Midjourney V6.1 demonstrates outstanding Aesthetic Quality (3.38), tying for first place, but performs relatively weakly in Prompt Following (3.36) and structural accuracy (3.31). Its strong visual appeal gives it a distinct advantage in Art and Film scenarios, making it ideal for applications focused on visual style exploration, creative art, atmospheric expression, and impactful visual presentation.
\end{itemize}
\begin{itemize}
    \item Ideogram 3.0: Mid-tier model with specialized strengths. Ideogram 3.0 also has balanced scores across three dimensions, though lower than top-tier models. Its distinct strength lies in text rendering, making it especially suitable for applications like logo design, posters, and text-image layout for office illustrations where textual clarity is essential. ELO analysis further supports this, showing that Ideogram 3.0 achieves higher win rates on prompts requiring explicit text rendering.
\end{itemize}
\begin{itemize}
    \item Flux 1.1 [pro]: Relatively weak performance. Flux 1.1 [pro] records the lowest ELO score (1,010) among the six models, with limited structural and Prompt Following capabilities, but comparatively strong cinematic realism, reflected in high Aesthetic Quality score (3.19) in Film.
\end{itemize}

\begin{table}[htbp]
\centering
\begin{adjustbox}{max width=0.99\textwidth}
\begin{tabular}{llcccccc}
\toprule
& & \textbf{Seedream 3.0} & \textbf{GPT-4o} & \textbf{Imagen 3} & \textbf{Ideogram 3.0} & \textbf{Flux 1.1 [pro]} & \textbf{Midjourney V6.1} \\
\midrule
\multirow{15}{*}{\textbf{Element}}
& Text                      & 0.82 & {\bf 0.95} & 0.57 & 0.88 & 0.50 & 0.48 \\
& Quantity                  & 0.69 & 0.65 & 0.60 & {\bf 0.74} & 0.69 & 0.55 \\
& Facial Expression         & 0.88 & {\bf 0.95} & 0.82 & 0.68 & 0.62 & 0.50 \\
& Texture                   & {\bf 0.93} & 0.88 & 0.81 & 0.72 & 0.72 & 0.81 \\
& Size                      & {\bf 0.87} & {\bf 0.87} & 0.82 & 0.71 & 0.76 & 0.80 \\
& Comparative Relation      & 0.69 & {\bf 0.71} & 0.59 & 0.61 & 0.67 & 0.42 \\
& Compositional Relation    & 0.76 & {\bf 0.80} & 0.74 & 0.58 & 0.51 & 0.51 \\
& Content Relation          & {\bf 0.83} & 0.79 & 0.79 & 0.65 & 0.62 & 0.58 \\
& Similarity Relation       & 0.65 & {\bf 0.73} & 0.64 & 0.63 & 0.59 & 0.54 \\
& Full-body Action          & 0.93 & {\bf 0.95} & 0.85 & 0.86 & 0.75 & 0.74 \\
& Partial Action            & 0.93 & 0.91 & 0.87 & 0.91 & 0.72 & 0.74 \\
& Animal Action             & 0.73 & {\bf 0.92} & 0.77 & 0.82 & 0.66 & 0.60 \\
& Contact Interaction       & 0.86 & {\bf 0.87} & 0.74 & 0.72 & 0.64 & 0.59 \\
& Non-contact Interaction   & {\bf 0.83} & 0.79 & 0.70 & 0.70 & 0.58 & 0.50 \\
& State                     & 0.89 & {\bf 0.90} & 0.87 & 0.72 & 0.81 & 0.76 \\
\midrule
\multirow{3}{*}{\textbf{Element Composition}}
& Multi-Entity Feature Matching & 0.58 & {\bf 0.81} & 0.41 & 0.35 & 0.33 & 0.12 \\
& Layout \& Typography           & 0.87 & {\bf 0.93} & 0.67 & 0.77 & 0.71 & 0.55 \\
& Anti-Realism                   & 0.66 & {\bf 0.95} & 0.64 & 0.66 & 0.51 & 0.44 \\
\midrule
\multirow{3}{*}{\textbf{Text Expression Form}}
& Negation               & 0.28 & {\bf 0.98} & 0.79 & 0.69 & 0.68 & 0.41 \\
& Pronoun Reference      & 0.68 & {\bf 1.00} & 0.61 & 0.86 & 0.52 & 0.43 \\
& Consistency            & 0.59 & {\bf 0.81} & 0.68 & 0.61 & 0.44 & 0.39 \\
\bottomrule
\end{tabular}
\end{adjustbox}
\caption{Evaluation results of mainstream T2I models on objective capability test points. Score ranges from 0--1; values in bold indicate the top-performing model for each test point.}
\label{tab:element_scores}
\end{table}

To further examine how current models perform across various capabilities, we conducted test points scoring, with the detailed evaluation methodology provided in Appendix~\ref{appendix:D}.
The experimental results (Table~\ref{tab:element_scores}) reveal that each model also exhibits divergent characteristics across test points. For instance, GPT-4o leads across multiple test points with pronounced strengths in semantics-related test points, including Text (0.95), Facial Expression (0.95), Element Composition (Multi-Entity Feature Matching 0.81, Layout and Typography 0.93, Anti-Realism 0.95), and Text Expression Form (Negation 0.98, Pronoun Reference 1.00, Consistency 0.81). In contrast, Seedream 3.0 demonstrates superiority in visual and action-related test points, such as Texture (0.93), Full-body and Hand Actions (both 0.93), and Interaction Actions (Contact 0.86, Non-contact 0.83).

Furthermore, test point analysis reveals that all models consistently achieve low scores on certain tasks, such as those involving complex relational reasoning (e.g., comparative and similarity relations), quantity, element composition, and text expression forms. Although mainstream T2I models have made notable progress, significant limitations persist, highlighting areas for future improvement.

\subsection{Joint analysis of ELO and MOS}
%By integrating the overall ELO rankings with dimension-specific MOS scores, we conducted a regression analysis to investigate how each dimension influences overall model performance.
%We user a multivariate logistic regression that uses individual ELO matchups to estimate how increasing in each dimension's MOS score affects the probability of winning. The MOS values for the three dimensions are first standardized because these dimensions exhibit different score dispersions. To account for potential inter-dimension correlations, we augment the regression with interaction terms as well as x·Δx terms. The effect of a one–sigma increment is not uniform across the 1--5 score range; for instance, a shift from 1 to 2 differs from a shift from 4 to 5. For expositional clarity, we present an averaged regresssion result across personas and scenarios in Table~\ref{tab:weights_persona} and Table~\ref{tab:Application}. Prompt Following emerges as the most important determinant of user satisfaction, as it directly reflects intent fulfillment. This is verified by different and scenarios. This aligns with intuitive expectations, since when an image does not accurately represent the core semantics of the prompt, strengths in other dimensions have limited practical significance.
We employ a multivariate logistic regression using individual ELO matchups to estimate how increases in each dimension's MOS score affect ELO win rate. Given the varying dispersion across the three dimensions, their MOS values are standardized before regression. The model further includes cross-variable interaction (such as $x_1 \cdot x_2$) and level-change interaction (such as $x \cdot \Delta x$) terms to account for potential interdependencies and non-linear effects. The one-standard-deviation increase is nonuniform across the 1--5 scale (e.g., a shift from 1 to 2 differs from 4 to 5). For clarity, we present only the aggregated one-standard-deviation increase at the dimension level in Table~\ref{tab:weights_persona} and Table~\ref{tab:experts_weights}, which respectively show the results for different personas and scenarios. It is worth noting that across all results, Prompt Following emerges as the most important determinant of user selection, as it directly reflects intent fulfillment. This observation aligns with intuitive expectations, since when an image does not accurately represent the core semantics of the prompt, strengths in other dimensions have limited practical significance.

%The fitting method uses a multivariate logistic regression that effectively leverages each individual ELO matchup:
%\begin{itemize}
%    \item Dependent variable (Y): overall preference outcome.
%    \item Independent variables (X): expert MOS score differences across the three metrics---Prompt Following, Structural Accuracy, and Aesthetic Quality.
%    \item To account for potential inter-metrics correlations, the model includes interaction (cross) terms.
%\end{itemize}

%\textbf{Dimension weight interpretation.} The three evaluation dimensions contribute differently to user satisfaction. Table~\ref{tab:weights_persona} shows the absolute increase in Elo match win rate when the MOS of a given dimension increases by one point (after standardization), holding the other dimensions constant.

%Prompt Following emerges as the most important determinant of user satisfaction, as it directly reflects intent fulfillment. This aligns with intuitive expectations, since when an image does not accurately represent the core semantics of the prompt, strengths in other dimensions have limited practical significance.

\begin{table}[!htbp]
\centering
\small
\setlength{\tabcolsep}{6pt}
\begin{adjustbox}{max width=0.9\textwidth}
\begin{tabularx}{\textwidth}{%
  L % Persona 列，左对齐
  C C C % 三个指标列，居中
}
\toprule
\multirow{2}{=}{\textbf{Persona}} 
& \multicolumn{3}{c}{\parbox{0.7\textwidth}{\centering The increment in absolute ELO match win rate (from a 50\% baseline) associated with a one-standard-deviation increase in standardized MOS score.}} \\
\cmidrule(lr){2-4}
& \textbf{Prompt Following} & \textbf{Structural Accuracy} & \textbf{Aesthetic Quality} \\
\midrule
General user & 12.5\% & 0.7\% & 7.1\% \\
Expert & 37.7\% & 14.2\% & 10.9\% \\
Designer & 26.0\% & 10.0\% & 22.1\% \\
\bottomrule
\end{tabularx}
\end{adjustbox}
\caption{Average evaluation dimension weights across different personas, including those in ELO public and expert mode, as well as the distinctive designer subgroup within both modes. Suppose two images A and B, have the same MOS scores in Prompt Following, Structural Accuracy, and
Aesthetic Quality. Consequently, both images would exhibit identical win rates (50\%) in ELO. If image A's Prompt Following score increases by one-standard-deviation relative to image B, its win rate rises to 62.5\% ($50\%+12.5\%$). In comparison, a one-standard-deviation increase in Structural Accuracy raises the win rate to just 50.7\% ($50\%+0.7\%$).}
\label{tab:weights_persona}
\end{table}
\begin{table}[!htbp]
\centering
\small
\setlength{\tabcolsep}{8pt}
\begin{adjustbox}{max width=0.9\textwidth}
\begin{tabular}{l c c c c}
\toprule
\textbf{Application Scenarios} & \textbf{ELO score of Midjourney V6.1} & \multicolumn{3}{c}{\textbf{Experts' weights}} \\
\cmidrule(lr){3-5}
& & Prompt Following & Structural Accuracy & Aesthetic Quality \\
\midrule
Art & \textcolor{red}{1003} & \textcolor{red}{32.7\%} & 12.5\% & 10.4\% \\
Entertainment & 964 & 35.8\% & 15.1\% & 10.9\% \\
Aesthetic Design & 931 & 38.6\% & 13.9\% & 10.0\% \\
Functional Design & 936 & 38.5\% & 18.6\% & 9.4\% \\
Film & 910 & 40.6\% & 11.3\% & 8.7\% \\
\bottomrule
\end{tabular}
\end{adjustbox}
\caption{Illustration of how dimension weights differ across application scenarios, using experts' weights as a representative example. The red-highlighted numbers indicate that Prompt Following receives a relatively lower weight in the art scenario, and that Midjourney V6.1 achieves the highest ELO score in this domain.}
\label{tab:experts_weights}
\end{table}

\textbf{Population preferences.} The weighting of evaluation dimensions varies among different users (in Table~\ref{tab:weights_persona}).
\begin{itemize}
    \item Experts exhibit a greater increment in win rate compared to general user across all three dimensions---Prompt Following ($\beta = 37.7\%$ vs. $12.5\%$), Structural Accuracy ($\beta = 14.2\%$ vs. $0.7\%$), and Aesthetic Quality ($\beta = 10.9\%$ vs. $7.1\%$). This further corroborates our earlier observation that experts are more sensitive to subtle differences.
    \item For general users, it is notable that the weight for Structural Accuracy is below 1\%, markedly lower than those for Prompt Following (12.5\%) and Aesthetic Quality (7.1\%). The average Structural Accuracy score for mainstream T2I models already exceeds 3, indicating robust performance. As a result, minor structural deviations such as errors in hands or small background objects have limited impact on general users' selection of superior images. In contrast, pronounced structural anomalies, such as limb errors resulting in MOS scores of 1 or 2, still strongly diminish overall user preference.
\end{itemize}

Additionally, cross-occupation analysis shows that designers from both general users and experts exhibit distinctive aesthetic sensitivity. Their aesthetic weight reaches 22.1\%, which is 3.1 times that of general users (7.1\%). This underscores the strong influence of professional background on selection preferences.

\begin{figure}[!htbp]
\centering
\includegraphics[width=0.9\linewidth]{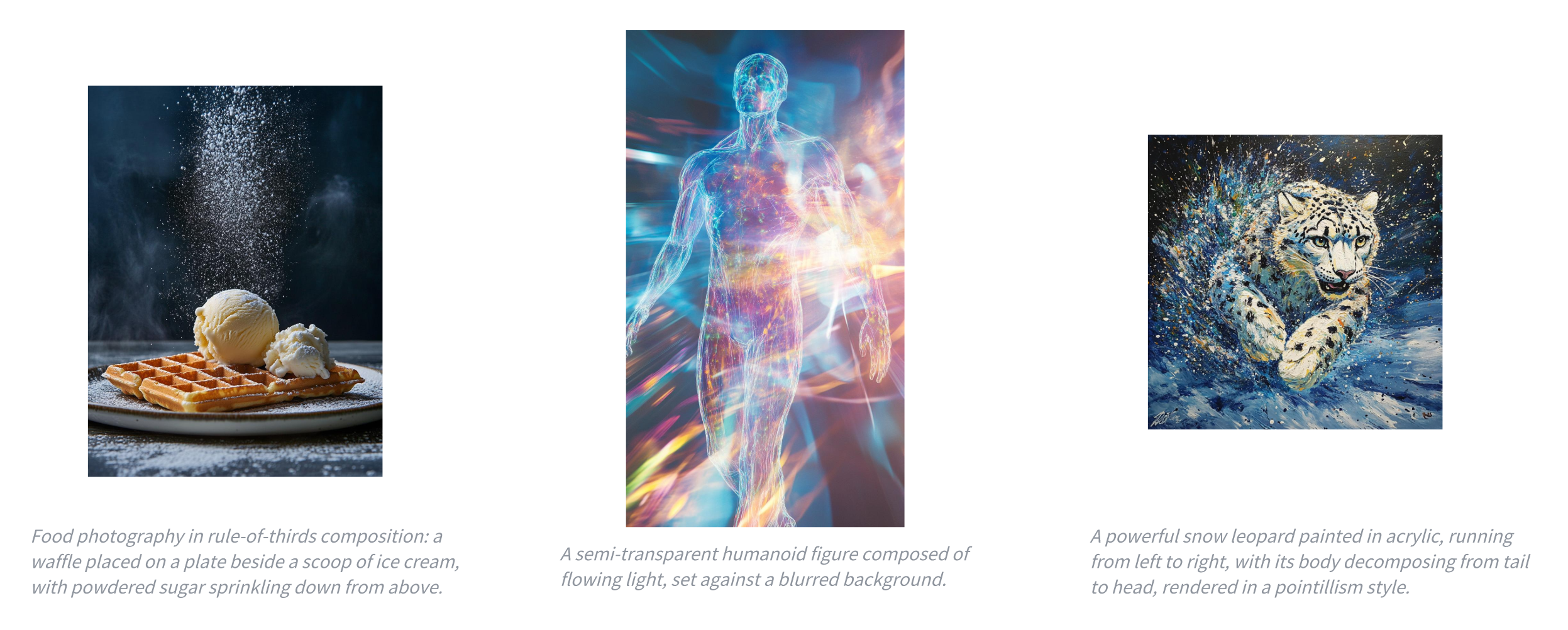}
\caption{\label{fig:mj_art}These prompts are selected from artistic scenarios, and  images are generated by Midjourney, which exhibit a particularly strong visual impact.}
\end{figure}

\textbf{Application-scenario preferences.} As shown in Table~\ref{tab:experts_weights}, the dimension weights vary across scenarios, with art being particularly distinctive in that Prompt Following receives the lowest weight among all five scenarios. This helps explain why Midjourney V6.1 ranks much higher in the art scenario---its strong visual impact partially offsets its weakness in Prompt Following, as demonstrated by the three examples in Fig.~\ref{fig:mj_art}.

%% file: 7_conclusion.tex
\section{Conclusion}
This paper presents the Magic Evaluation Framework (MEF), a systematic and practical approach for evaluating T2I generation models. The core contributions of the framework encompass two aspects:

\begin{itemize}
    \item \textbf{Evaluation methodology.} MEF ensures thorough assessment from both application and diagnostic perspectives by integrating an ELO-based ranking system with a multi-dimensional, detailed MOS scoring system. ELO captures overall user satisfaction and enables robust model ranking. The MOS scoring complements this by providing fine-grained diagnostic analysis through expert reviews to identify issues in the dimensions of Prompt Following, Structural Accuracy, and Aesthetic Quality. Joint analysis of ELO and MOS further quantifies the impact of each dimension on model performance, guiding iteration priorities. Reliability is ensured through a comprehensive quality control system aligned with the evaluation framework, including expert admission, anchor-item monitoring, statistical testing, and other measures.
    
    \item \textbf{Evaluation benchmark.} Guided by a label-oriented design strategy, we constructed the Magic-Bench-377 benchmark, which is built upon a multi-level taxonomy. This benchmark comprehensively covers three core capability categories---Element, Element Composition, and Text Expression Form---and five major application scenarios, ensuring that the evaluation addresses both the boundaries of model capabilities and the enhancement of user experience. Furthermore, the hierarchical taxonomy enables convenient extension and refinement of labels, supporting fine-grained test point conclusions and facilitating ongoing adaptation to emerging evaluation needs. 
\end{itemize}
Experimental results indicate that while mainstream models have made significant progress in foundational capabilities, top-tier models still exhibit notable deficiencies, particularly in complex test points. Prompt Following remains the most critical evaluation dimension across user groups and scenarios, while the importance of other dimensions varies. Collectively, these findings provide actionable guidance for model iteration, supporting targeted enhancements in areas with the greatest performance gaps and potential impact.

%% file: 8_future_work.tex
\section{Limitations $\&$ Future work }
While the MEF provides a systematic framework for evaluating T2I models, several limitations remain that should be addressed in future work:

\begin{itemize}
    \item \textbf{Evaluation of long-tail and combinatorial test factors.}
        Constrained by cost and efficiency, the current Magic-Bench-377 benchmark includes only 377 test cases, leaving limited coverage of long-tail scenarios and complex combinatorial tasks. Therefore, in our practical evaluation process, we incorporated dedicated assessment sets, which include both manual and automated evaluations, to address these gaps. We encourage peers who utilize the MEF to expand their assessments as well, to obtain a more comprehensive understanding of model capabilities.
    \item \textbf{Automation to improve evaluation efficiency.}
        As models advance and scenarios diversify, human-based evaluation faces increasing efficiency bottlenecks. Future research will focus on exploring automated evaluation methods that meet accuracy standards, aiming to gradually replace manual assessments and thereby improve efficiency without compromising reliability.
    \item \textbf{Extending to multi-modal capability evaluation.}
        The current Magic-Bench-377 design is primarily focused on single-image text-to-image generation tasks and does not yet cover scenarios such as multi-modal conditioned generation, or tasks that demand multi-modal outputs. However, user demand in these scenarios is considerable, and as model capabilities evolve, support and performance on these tasks will become central to evaluation. Accordingly, future work will extend both the benchmark and methodology to multi-modal capability scenarios.
\end{itemize}

%% file: appendix.tex
\section{Label Coverage Statistics}\label{appendix:A}
To ensure consistency, we adopt the proposed labeling framework to compare label coverage across existing benchmarks. Since specialized benchmarks such as Commonsense and WISE target only a single dimension of objective capability, our comparison focuses on comprehensive benchmarks, including PartiPrompt~\cite{Yu2022}, DrawBench~\cite{Saharia2022}, and DEsignBench~\cite{Lin2023} (which additionally incorporates application scenarios). The comparison is conducted by mapping the skill categories defined in these benchmarks onto our labeling framework.
\clearpage
\section{Detailed Definitions of Taxonomy and Example Case}\label{appendix:B}
\begin{table}[htbp]
\centering
\small
\setlength{\tabcolsep}{8pt}
\begin{tabularx}{\textwidth}{%
  >{\raggedright\arraybackslash}p{3cm} % Capability Category，固定宽度，左对齐
  >{\raggedright\arraybackslash}p{3cm} % Capability Item，固定宽度，左对齐
  L % Definition，自动换行左对齐
  L % Example，自动换行左对齐
}
\toprule
\textbf{Capability Category} & \textbf{Capability Item} & \textbf{Definition} & \textbf{Example} \\
\midrule
\multirow{3}{=}{Element Granularity} 
& Entity & Semantic units referring to entities such as people, animals, scenes, costumes, and decorations, including real-world and virtual entities, man-made objects, and natural elements. 
& \textit{A little \textbf{girl} in a \textbf{dress} walking her pet \textbf{dog} in a \textbf{park} encounters a \textbf{robot}.} \\
& Entity Description & Semantic units describing the quantity, attributes, forms, states, or relationships of entities. 
& \textbf{\textit{Three}} children dressed in \textbf{\textit{jeans}} are \textbf{\textit{playing jump rope}}, while \textbf{\textit{several pink}} roses are in full bloom beside them. \\
& Image Description & Semantic units that describe visual elements of a scene, including style, aesthetics, and artistic knowledge. 
& \textit{Light mist rises from a mountain valley, depicted in an \textbf{ink wash painting} with \textbf{minimalist composition}.} \\
\midrule
\multirow{3}{=}{Element Composition} 
& Anti-Realism & Combinations that contradict real-world cognition or physical laws. 
& \textbf{\textit{A ceramic cup floating in the air.}} \\
& Layout \& Typography & Descriptions of spatial or positional relationships among images, text, or symbols. 
& \textit{A poster with the word "TOPIC" at the top and "Love \& Peace" at the bottom.} \\
& Multi-Entity Feature Matching & Multiple entities of the same type with distinct attribute values. 
& \textbf{\textit{Three purple gemstones}} and \textbf{\textit{one pink gemstone}}. \\
\midrule
\multirow{3}{=}{Text Expression Form} 
& Negation & Negative expressions such as "no", "without" or "does not". 
& \textit{A fish tank \textbf{without} any fish.} \\
& Pronoun Reference & Pronouns (he, she, it, they) referring back to entities mentioned earlier in the text, requiring the model to resolve co-reference. 
& \textit{The bear lies on the ground, the cub lies beside \textbf{it}.} \\
& Consistency & Multiple entities of the same type sharing the same attribute. 
& \textit{A group of people \textbf{all} wearing sunglasses, sunbathing on the beach.} \\
\bottomrule
\end{tabularx}
\caption{Definitions and Example Cases of Model Capability Labels. In the example cases, the \textbf{bolded text} indicates the elements that are assigned the corresponding category labels.}
\label{tab:capability_labels}
\end{table}

\begin{table}[htbp]
\centering
\small
\setlength{\tabcolsep}{8pt}
\begin{tabularx}{\textwidth}{%
  >{\raggedright\arraybackslash}p{3cm} % Application Scenario，固定宽度
  L % Definition，自动换行左对齐
  L % Example，自动换行左对齐
}
\toprule
\textbf{Application Scenario} & \textbf{Definition} & \textbf{Example} \\
\midrule
Film & Focuses on user needs for story-driven content creation, such as storyboards, cinematic scenes, or animated sequences. Models are expected to understand narrative details and generate scenes with coherent environments and character interactions.
& \textit{A terrifying movie scene: in a dark alley at night, a tall man holding a sharp knife stands in front of a couple, with the girl hiding behind her boyfriend.} \\
Art & Focuses on user needs for high-level artistic creation, requiring models to generate outputs aligned with artistic styles, aesthetics, and visual imagination, such as oil painting, watercolor, sketching, or abstract expression.
& \textit{An artwork combining charcoal drawing and red ink, exploring the theme of depression. In the foreground, crisscrossing barbed wire is depicted, while at the center of the composition a curled-up human figure is enveloped in darkness, evoking a strong sense of isolation.} \\
Entertainment & Focuses on user needs for casual, creative and entertaining content, often reflecting internet culture (e.g., memes, emojis, or playful illustrations). The goal is to stimulate fun, amusement, or humor.
& \textit{A series of emoji-style designs: Expression 1: A monkey grinning widely, with the caption "Happy". Expression 2: A monkey wearing sunglasses, with the caption "Cool". Expression 3: A monkey holding a flower with a shy expression, with the caption "Shy". Expression 4: A monkey showing a surprised expression, with the caption "Surprise".} \\
Aesthetic Design & Focuses on model use as a visual tool in professional design contexts, such as poster design, logo design, product design, etc. Models are expected to provide visually appealing outputs with high aesthetic quality.
& \textit{A fashion show poster featuring models in elegant dresses, with "MagicArena" in bold letters, a futuristic tech magazine cover titled "AI FASHION".} \\
Functional Design & Focuses on user needs for practical work and learning materials, such as teaching slides, product manuals, or office diagrams. Outputs emphasize clarity, conciseness and informativeness.
& \textit{A PowerPoint slide on "Curriculum, Practice, Activities, Collaboration," with each point illustrated by an icon of a leaf.} \\
\bottomrule
\end{tabularx}
\caption{Definitions and Example Cases of Application Scenario Labels.}
\label{tab:application_scenario_labels}
\end{table}

\FloatBarrier
\clearpage
\section{Rules for Constructing Evaluation Prompts}\label{appendix:C}
\begin{figure}[!htbp]
\centering
\includegraphics[width=0.85\linewidth]{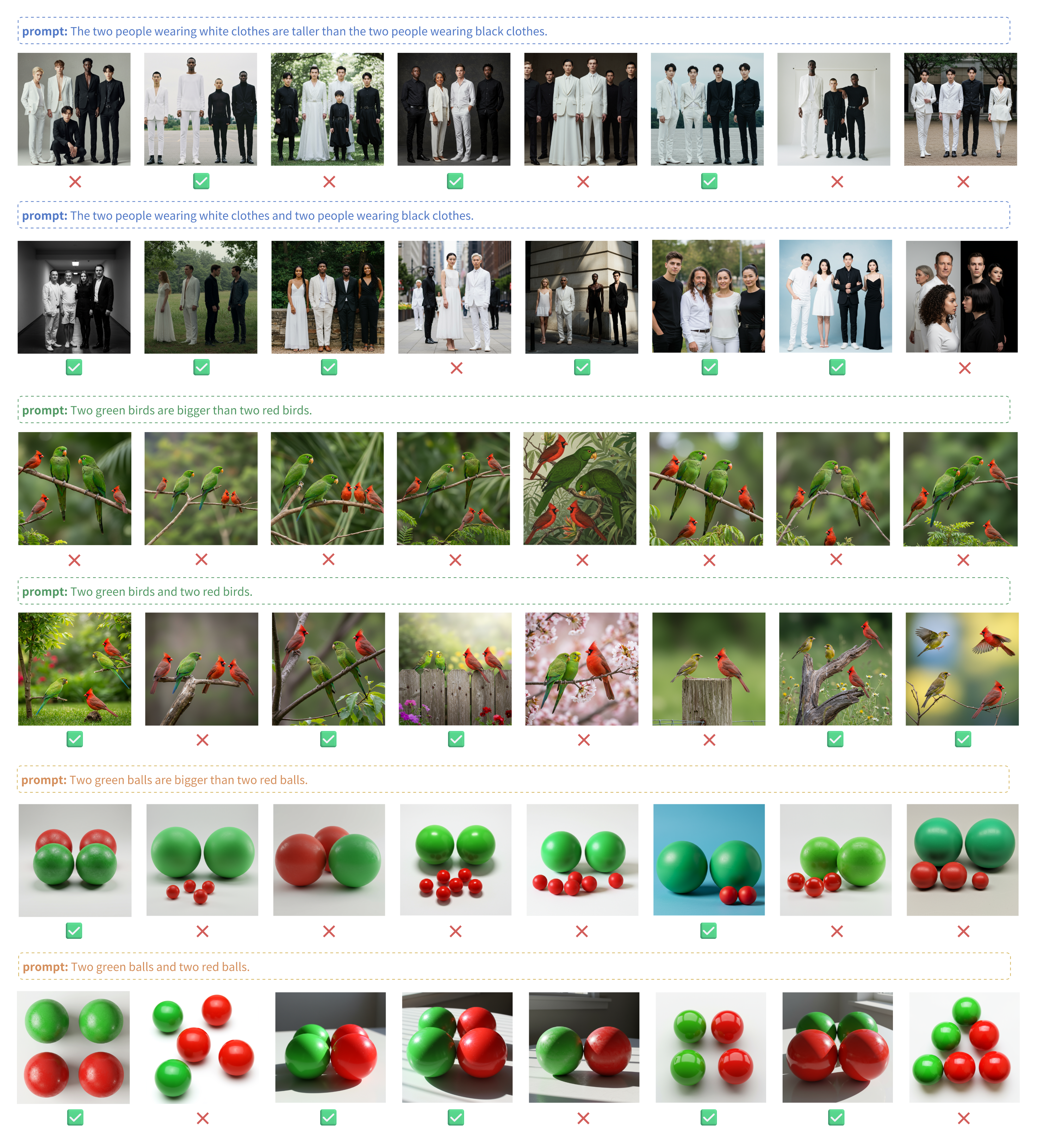}
\caption{\label{fig:frog}Combining test points may be related to a decrease in model performance. In all three examples, quantity, size, and color relationships are tested together. Excluding the size relationship results in a better performance on the quantity. }
\end{figure}

\begin{itemize}
    \item \textbf{Clarity and Visualizability.} Prompts should be concise, explicit and easy to visualize, while avoiding vague or non-visualizable descriptions. For example, the prompt ``A humanoid bird creature with golden wings spread wide, as if guarding a lush green mountain'' contains the phrase ``as if guarding a lush green mountain'', which is difficult to visualize, thereby limiting the ability to distinguish model performance.
    \item \textbf{Neutrality and Fairness.} Only generic characters and themes should be used when constructing evaluation data. Descriptions involving regional specificity, references to celebrities, or copyrighted characters must be avoided to ensure fair evaluation across all models. For instance, the prompt ``A Double Ninth Festival poster: generate a background image of climbing to honor ancestors, with a festival poem written on the right side'' would disadvantage models not trained on Chinese cultural contexts.
    \item \textbf{Diversity within Each Label.} Multiple prompts under the same label should remain diverse, rather than relying on identical elements or expressions. For example, under the texture label, prompts should cover varied textures such as metal, glass and silk, instead of repeatedly using only one or two textures.
    \item \textbf{Multiple Test Points per Prompt.} Prompts should reflect real user requirement by composing multiple capabilities within a single description, rather than isolating capabilities one by one. For instance, a prompt may specify quantity, color, and spatial relations together in Fig.~\ref{fig:frog}.
\end{itemize}
\clearpage
\section{Test Point Evaluation Method}\label{appendix:D}
To enable a more fine-grained analysis of model performance across different objective capabilities, each item in Magic-Bench-377 was decomposed according to its annotated capability labels, resulting in multiple evaluation samples that each test only a single capability. During scoring, evaluators were required to determine whether the generated image satisfied the specific capability requirement. A score of 1 was assigned if the requirement was fully met, and 0 otherwise. This evaluation focused exclusively on the Prompt Following dimension and did not assess Structural Accuracy or Aesthetic Quality.
\clearpage
\section{ELO and MOS Scoring Standard}\label{appendix:E}
The scoring rules for both ELO scores and multi-dimensional MOS scores are introduced below.

\begin{figure}[!htbp]
\centering
\includegraphics[width=0.4\linewidth]{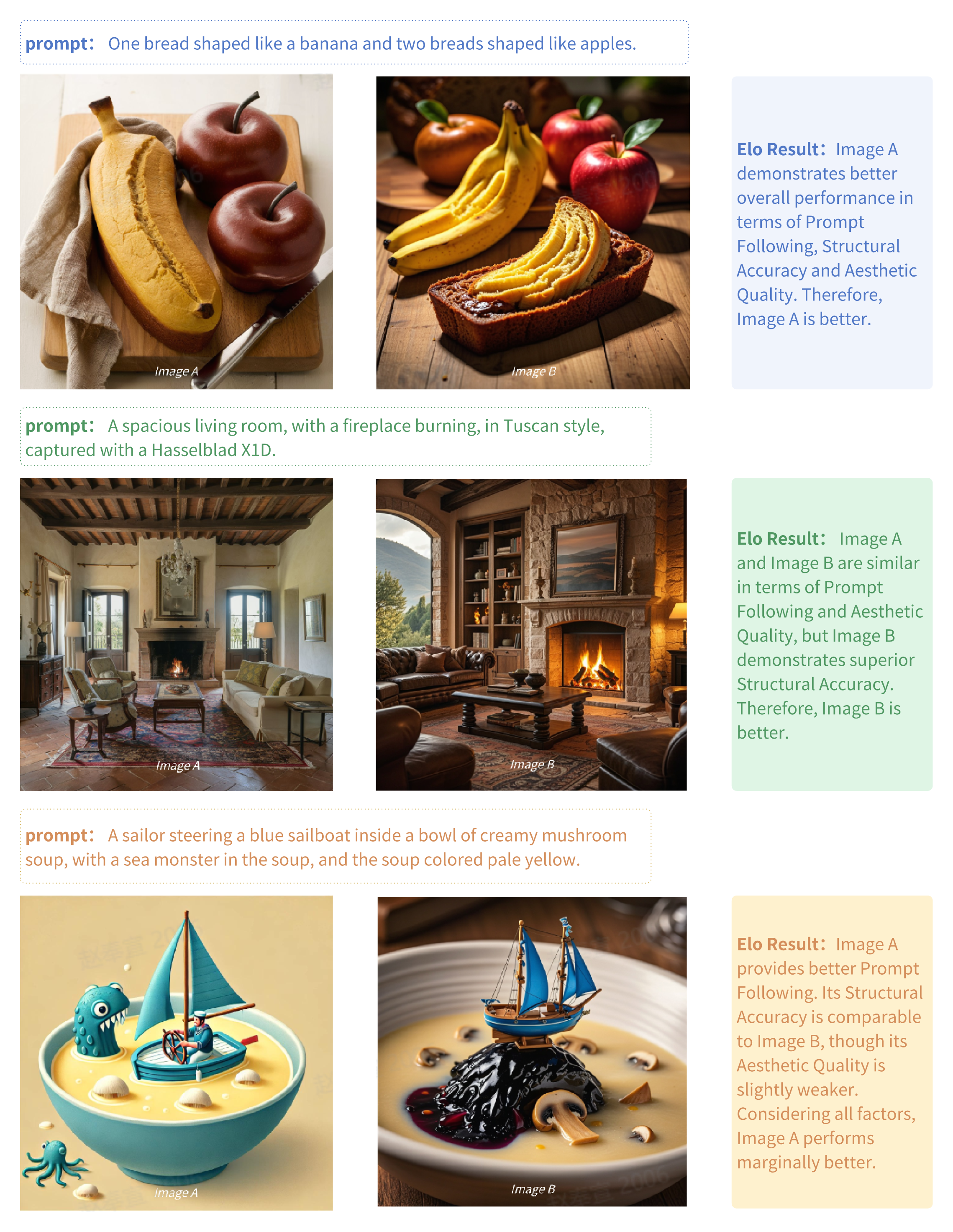}
\caption{\label{fig:elo_sd}Representative case examples illustrating the ELO evaluation rules.}
\end{figure}

\textbf{ELO Score.} The ELO score (in Fig.~\ref{fig:elo_sd}) primarily measures the overall degree to which a generated image satisfies the input text prompt, requiring a holistic consideration of performance across the three dimensions: Prompt Following, Structural Accuracy and Aesthetic Quality. If Image A performs no worse than Image B across all three dimensions, then Image A is judged superior. However, when Image A and Image B each show strengths and weaknesses across different dimensions, the evaluator must make a comprehensive judgment based on the relative weights of the dimensions and the severity of the observed issues.

\begin{figure}[!htbp]
\centering
\includegraphics[width=0.64\linewidth]{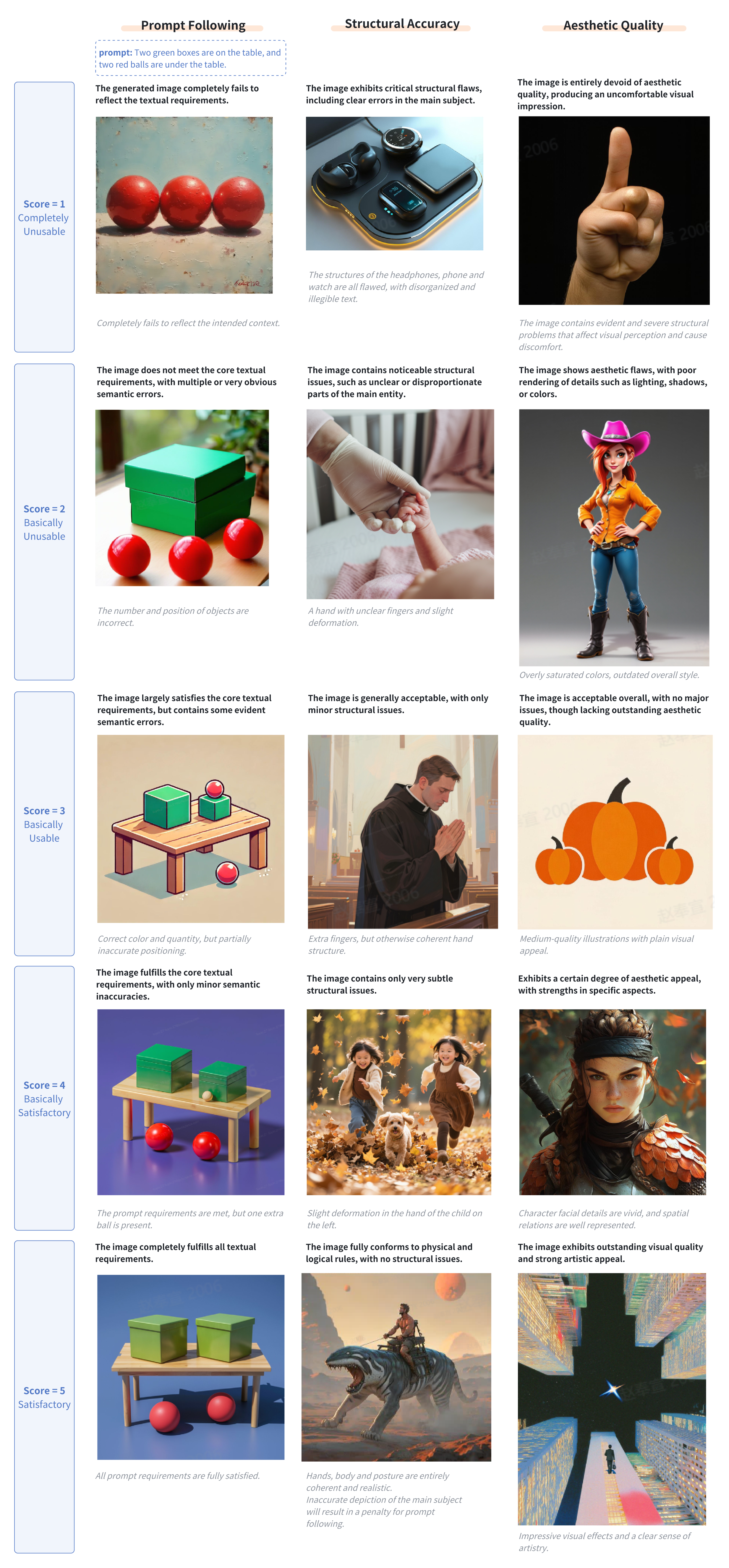}
\caption{\label{fig:mos_sd}Evaluation rules for MOS scoring and illustrative case examples.}
\end{figure}

\textbf{MOS Score. }The MOS scoring system evaluates model performance across the three core dimensions: Prompt Following, Structural Accuracy and Aesthetic Quality. Each dimension is rated on a 1--5 scale, with the meaning of each score level and the corresponding scoring guidelines detailed in Fig.~\ref{fig:mos_sd}.

\clearpage
\section{Scoring Interface}
\begin{figure}[H]
\centering
\includegraphics[width=0.9\linewidth]{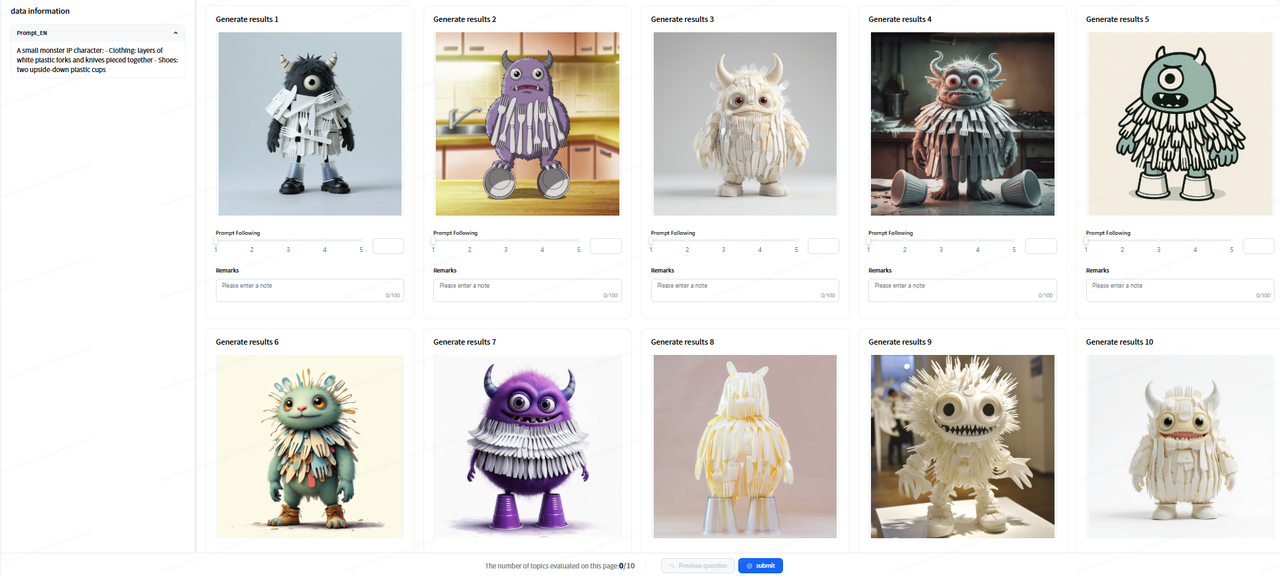}
\caption{\label{fig:frog1}MOS scoring interface.}
\end{figure}
\begin{figure}[H]
\centering
\includegraphics[width=0.9\linewidth]{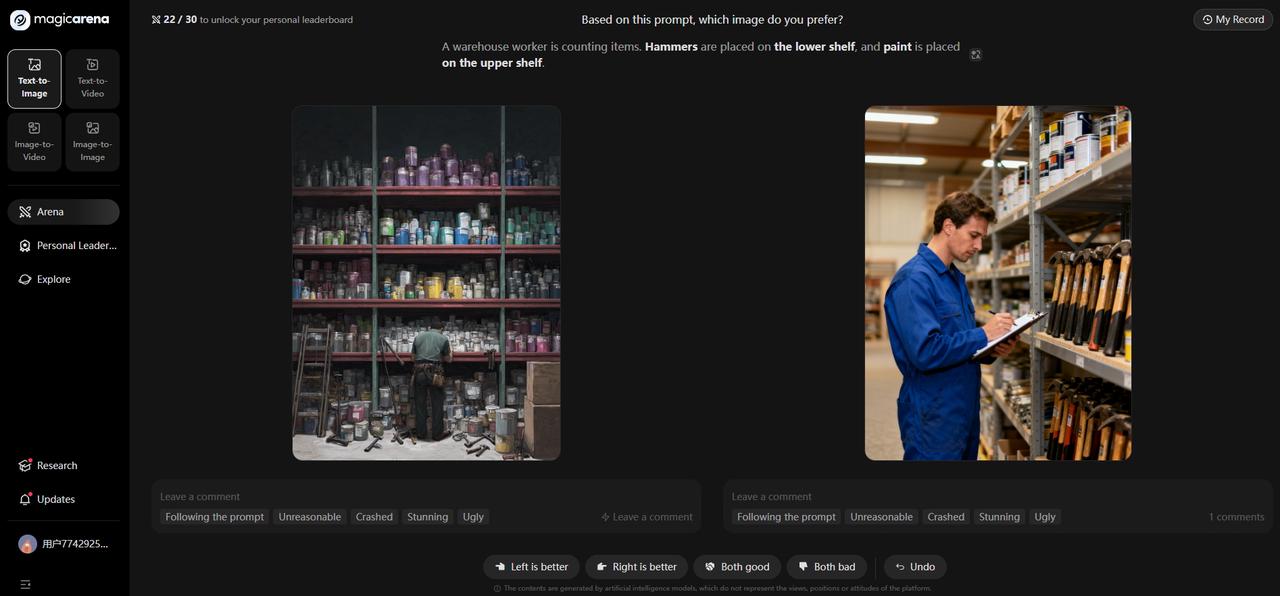}
\caption{\label{fig:frog2}ELO interface.}
\end{figure}

\clearpage
\section*{Contributions}
\textbf{\textcolor{black}{Core Contributor}}\\
Xiaojing Dong, Weilin Huang, Liang Li, Yiying Li, Shu Liu, Tongtong Ou, Shuang Ouyang, Yu Tian, Fengxuan Zhao

\vspace{1em} % 空一行

\textbf{\textcolor{black}{Contributor}}\\
Xinqi Cheng, Jing Cui, Jing Fang, Meng Guo, Bibo He, Hongwei Kou, Hao Li, Yameng Li, Botao Liu, Yin Liu, Liyue Wang, Junkai Wang, Qingyi Wang, Ruolan Wu, Jianchao Yang, Linkai Zeng, Zhonghua Zhai, Xiaoyin Zhang, Yongjian Zhang, Huating Zhao, Feilong Zuo
\section*{Acknowledgments}
We thank all colleagues at ByteDance for their support of this project.